\documentclass[11pt]{article}

\usepackage[letterpaper,margin=1in]{geometry}

\usepackage[utf8]{inputenc}
\usepackage[T1]{fontenc}
\usepackage{lmodern}        % scalable Latin Modern (no bitmap fonts)
\usepackage{microtype}      % subliminal kerning and protrusion

\usepackage{amsmath}
\usepackage{amsfonts}
\usepackage{amssymb}
\usepackage{nicefrac}

\usepackage{graphicx}
\usepackage{booktabs}
\usepackage[font=small,labelfont=bf,skip=6pt]{caption}
\usepackage{placeins}

\usepackage{xcolor}
\definecolor{linkcol}{HTML}{1A3A8F}
\definecolor{citecol}{HTML}{1A3A8F}
\definecolor{urlcol} {HTML}{1A3A8F}
\usepackage{url}
\usepackage[
  colorlinks=true,
  linkcolor=linkcol,
  citecolor=citecol,
  urlcolor=urlcol,
  pdftitle={AI Rater Discrimination Depends on Scoring Protocol in Complex Clinical Decision-Making},
  pdfauthor={Sangwon Baek, Kyu Yeon Hur, Kyunga Kim},
  pdfborder={0 0 0},
  breaklinks=true
]{hyperref}

\title{AI Rater Discrimination Depends on Scoring Protocol \\
       in Complex Clinical Decision-Making}

\author{%
  \textbf{Sangwon Baek} \\
  Asclep Korea Inc., Changwon, Republic of Korea \\
  Center for Data Science, New York University, New York, USA \\
  \texttt{baeksw98@gmail.com} \\
  \and
  \textbf{Kyu Yeon Hur}\thanks{Drs.\ Kyunga Kim and Kyu Yeon Hur contributed equally as co-corresponding authors.} \\
  Division of Endocrinology and Metabolism, Department of Medicine, \\
  Samsung Medical Center, Sungkyunkwan University School of Medicine, Seoul, Korea \\
  \and
  \textbf{Kyunga Kim}\footnotemark[1] \\
  Biomedical Statistics Center, Samsung Medical Center; \\
  Department of Digital Health, SAIHST; \\
  Department of Data Convergence \& Future Medicine, \\
  Sungkyunkwan University, Seoul, Republic of Korea \\
  \texttt{kyunga.j.kim@gmail.com}
}

\date{}

\begin{document}

\maketitle

\begin{abstract}
Clinical AI evaluation increasingly delegates scoring to large language models (LLMs) acting as AI raters, yet their scoring behavior across evaluation conditions has not been quantitatively characterized. We address this gap through a factorial study of AI rater behavior in adult type 2 diabetes (T2D) pharmacotherapy at 12-month outpatient follow-up, a clinical task involving complex decision-making operationalized across seven evaluation questions. Four open-source LLMs served simultaneously as clinical decision support system (CDSS) models and AI raters. Each CDSS output was scored under two scoring protocols: a rubric-anchored Gold Rubric (GR) protocol incorporating  a patient-specific rubric, and a rubric-free Non Gold Rubric (Non-GR) protocol. Linear mixed effects models crossed the scoring protocol factor with five design factors---CDSS model, CDSS prompt configuration (document-referenced generation [DRG] vs.\ Baseline), rater model, prompt character, and prompt type---and estimated main effects together with their protocol interactions. Across all questions, AI raters yielded consistently higher scores within a very narrow range (74--78 points on average) under Non-GR compared to those under GR (7.69 to 49.64 points lower mean scores; 1.68 to 3.67 times wider interquartile ranges). Within each question, GR amplified the AI rater's discrimination between DRG and Baseline CDSS outputs by factors of 1.76 to 5.10, while also revealing substantial behavioral variation across rater models that Non-GR suppressed. These findings support rubric anchoring as the scoring protocol that preserves discriminative power in clinical AI evaluation; rubric-free scoring cannot substitute when questions require patient-specific or jurisdiction-specific criteria that rater models cannot infer from parametric knowledge alone.
\end{abstract}

%==================================================================
\section{Introduction}
\label{sec:intro}

Clinical AI evaluation has begun to incorporate large language models (LLMs) as AI raters that assign scores to clinical outputs produced by other AI systems at scale. This shift reflects two limits of human-based evaluation that have persisted across clinical domains. Human expert raters disagree with one another on the same case, producing interobserver variability documented in systematic reviews [1, 2]. Expert review also struggles to keep pace with the volume of AI-generated clinical outputs, and the fatigue that accompanies extended review correlates with measurable increases in clinical error [3]. Large-scale clinical benchmarks such as MedHELM [4] and HealthBench [5] have already adopted LLM juries as scoring mechanisms, signaling that AI rater--based evaluation is increasingly shaping how clinical AI systems are assessed.

Whether the trajectory of general-domain LLM-as-a-judge evaluation transfers cleanly to the clinical setting requires careful examination. The clinical setting is safety-critical, which places corresponding demands on any evaluator. A scoring error in clinical evaluation can misrepresent the safety or appropriateness of a treatment recommendation, and even small deviations in scoring can shift the conclusion drawn from the evaluation. The evaluator must therefore not merely approximate clinical judgment; its scoring behavior should be quantifiable and reportable to a degree that supports informed deployment. AI rater adoption may bring substantial value by enabling systematic, large-scale evaluation, but this benefit must come with a corresponding responsibility: AI raters should function as reliable evaluators if they are to be used for assessing clinical tools.

Meeting this responsibility requires more than demonstrating that AI raters produce reasonable-looking scores. It also requires knowing how those scores arise---under which conditions they remain stable, and under which they shift in ways that depend on prompt design rather than on the clinical content being scored. Such understanding becomes more pressing because LLMs are known to hallucinate---to generate outputs that read as plausible but contradict the underlying facts [6, 7]---and the same scoring-side instability has been documented directly in rater behavior, where models exhibit self-preference [8], sycophancy [9], and systematic sensitivity to prompt format [10]. Hallucination on the generative side and instability on the evaluative side share a common signature: outputs that look adequate but reflect prompt-driven properties rather than the content being judged. This pattern has not been eliminated under current architectures, but it can be characterized, quantified, and reported.

To perform this characterization in clinical AI evaluation, we conducted a factorial study of how AI raters behave across a range of evaluation conditions in type~2 diabetes (T2D) pharmacotherapy, a clinical task that involves complex decision-making over multiple drug classes. The study contrasts two scoring protocols drawn from current clinical evaluation practice: a Gold Rubric (GR) protocol, in which the rater receives a patient-specific rubric enumerating the required decision elements, and a Non Gold Rubric (Non-GR) protocol, in which the rater receives no rubric and scores from its own clinical knowledge. With this protocol contrast as the primary axis, we systematically vary clinical decision support system (CDSS) model, CDSS prompt configuration, rater model, prompt character, and prompt type, so that the contribution of each design choice can be estimated alongside its interaction with the protocol on the same CDSS outputs.

Our contributions are as follows:
\begin{itemize}
    \item We characterize how AI rater scoring behavior differs between the GR and Non-GR protocols across the seven evaluation questions (Q1--Q7), quantifying the difference in score distribution between the two protocols and how each component of the evaluation prompt contributes to scoring variance.
    \item We quantify whether the two protocols produce different score gaps between higher-quality and lower-quality CDSS outputs, characterizing how the choice of protocol affects the AI rater's ability to distinguish CDSS output quality.
    \item We examine how each individual rater model scores under the two protocols, comparing scoring tendencies across the four raters, the stability of each rater's scores across repeated runs, and each rater's scores on its own outputs versus outputs from other models.
\end{itemize}

%==================================================================
\section{Related Work}
\label{sec:related}

\paragraph{The emergence and validation of LLM-as-a-judge.} The LLM-as-a-judge paradigm developed as a scalable alternative to human evaluation as the volume of LLM outputs increased. Zheng et al.\ [11] formalized this approach with MT-Bench and Chatbot Arena and reported that LLM judges (such as GPT-4) reach high agreement with human experts. Subsequent work shifted toward characterizing systematic failure modes. Ye et al.\ [12] expanded the bias taxonomy across multiple models, and Koo et al.\ [13] reported bias indications across a range of LLMs that did not diminish with model scale. Zhu et al.\ [14] applied variance decomposition to general-domain evaluation and reported that systematic judge bias accounts for a substantial share of scoring variance at default operating points. The paradigm has therefore reached an empirical position in which LLM judges align with human raters at levels considered sufficient for practical use, while their scoring behavior is shaped by systematic biases that operate independently of model capability.

\paragraph{Characterizing evaluator biases.} Following the paradigm's validation, research has examined how evaluator biases operate within the LLM-as-a-judge framework. Wang et al.\ [15] and Li et al.\ [16] characterized position bias as a frequently replicated effect, with mitigation through paired comparisons in swapped positions now widely used. Panickssery et al.\ [8] reported self-preference mediated by perplexity, and Sharma et al.\ [9] reported sycophancy that increases when user preferences are disclosed. These findings indicate that a rater model cannot be assumed to score its own outputs neutrally. Sclar et al.\ [10], Nasser [17], Greenblatt et al.\ [18], and Serapio-Garc\'ia et al.\ [19] reported that rater models carry scoring dispositions that vary systematically between models, remain stable over time, and depend on prompt format to a degree that allows the rater model to be identified from its scoring patterns. Across this body of work, each bias has been studied in isolation within a fixed scoring protocol. Whether the scoring protocol itself---the structure of what the rater receives at the moment of evaluation---changes how these biases manifest on the same outputs has not been examined.

\paragraph{LLM-as-a-judge in clinical evaluation.} The paradigm has recently extended into safety-critical clinical domains, where rubric-anchored scoring has been used in several studies. MedHELM [4] reported a rubric-anchored clinical evaluation in which the LLM jury reached agreement with clinician ratings within the range observed among clinicians. HealthBench [5] reported a rubric framework with scoring criteria generated by physicians across a range of specialties. In medical education, Geathers et al.\ [20] reported agreement with expert scorers on OSCE evaluation. Across these clinical settings, AI scorers have been reported to assign higher scores than human evaluators; this leniency pattern coexists with verbosity bias documented in general-domain preference labeling [21], although the mechanistic link between the two has not been tested. Despite the use of rubric-anchored protocols across these studies, no clinical evaluation has tested whether the rubric itself---rather than the CDSS output being scored---accounts for the behavior its authors attribute to the AI rater.

\paragraph{Remaining gaps in the field.} Three gaps remain. In the general-domain LLM-as-a-judge literature, bias characterization has examined individual biases in isolation but has not comprehensively quantified diverse biases induced by multiple design choices. In clinical LLM evaluation, both rubric-anchored and rubric-free scoring approaches are in active use, yet no study has isolated the contribution of the rubric itself relative to a rubric-free alternative applied to the same CDSS outputs. At the intersection, no study has quantified how the scoring protocol---rubric-anchored versus rubric-free---interacts with rater model identity and rater-side prompt factors in shaping the observed score. These gaps converge on a single question: when the same CDSS output is scored by the same AI rater under two different scoring protocols, how does the rater's behavior change, and which design factors moderate that change most strongly?

%==================================================================
\section{Methodology}
\label{sec:methods}

\subsection{Overview}
\label{sec:methods-overview}

Our study applies a factorial experimental design to characterize how AI raters behave when evaluating CDSS outputs in T2D pharmacotherapy. The design isolates four sources of variation in AI rater scoring behavior: a binary protocol contrast between a Gold Rubric (GR) protocol and a Non Gold Rubric (Non-GR) protocol, a dual-role configuration in which the same four open-source LLMs serve as both CDSS models and AI raters, a CDSS prompt configuration factor (document-referenced generation [DRG] vs.\ Baseline) that produces CDSS outputs of two different quality levels, and a set of rater-side prompt factors crossed with the protocol. A glossary of all study-specific terms is provided in Appendix~\ref{app:glossary}.

\subsection{Protocol Manipulation}
\label{sec:methods-protocol}

The protocol factor contrasts two scoring configurations drawn from current clinical evaluation practice. The two configurations differ in the information the AI rater receives and the way the 0--100 score is produced.

\textbf{GR protocol.} Each patient is paired with a patient-specific rubric enumerating the required decision elements for the seven evaluation questions (Q1--Q7). The rubric was constructed by direct reference to the American Diabetes Association (ADA) Standards of Care in Diabetes [22] and validated by three senior board-certified diabetologists prior to use. The AI rater receives the rubric together with the CDSS output and scores each rubric element.

\textbf{Non-GR protocol.} The AI rater receives the CDSS output without any rubric and scores from its own internal clinical knowledge.

Both protocols are applied to every cell of the factorial design---the same CDSS output is scored once under GR and once under Non-GR---so within-cell scoring differences identify the protocol main effect and its interactions with the other design factors. The two protocols also differ in elicitation format (rubric-element aggregation under GR vs.\ direct 0--100 scoring under Non-GR); full mathematical specification is provided in Appendix~\ref{app:elicit}.

\subsection{Clinical Evaluation Task}
\label{sec:methods-task}

T2D pharmacotherapy was selected because it represents one of the most common settings for chronic-disease decision support, where the prescribing decision involves multiple drug types with distinct mechanisms and patient-specific safety constraints. The evaluation task scores AI-generated treatment recommendations for adult patients with T2D at 12-month outpatient follow-up [22]. The clinical domain is operationalized through seven evaluation questions, defined by expert consensus with senior board-certified diabetologists to cover the clinical decision flow, and applied to sixteen synthetic patients that span clinical variability across comorbidity profile, metabolic status, and anthropometric category. Full evaluation-question definitions, patient construction, and case-selection logic are provided in Appendix~\ref{app:cases}.

\subsection{Dual-Role Model Design}
\label{sec:methods-dualrole}

Four open-source LLMs form a shared pool that serves in two roles: each model acts as both a CDSS generator and an AI rater. This setup is referred to throughout the paper as the dual-role design, indicating that the same pool of models occupies both the generation and the scoring sides of the evaluation. The four models span architectural diversity in parameter scale, attention mechanism, and sparsity structure; full architectural specifications, Hugging Face repository references, and inference configurations are provided in Appendix~\ref{app:models}.

The selection was constrained to open-source models because clinical AI evaluation pipelines that ingest identifiable patient information typically remain within institutional infrastructure under data-protection regulations. Holding the model pool constant between the two roles ensures that observed cross-model differences in scoring behavior reflect rater-side conditions rather than differences in the underlying model pool.

\subsection{CDSS Prompt Configuration}
\label{sec:methods-cdss-prompt}

Two CDSS prompt configurations produce CDSS outputs of two different quality levels. Under the Baseline prompt, the CDSS receives the patient case alone and generates its output directly from its parametric clinical knowledge. Under the DRG prompt, the CDSS receives the patient case together with curated reference documentation prepared in advance and uses the documentation as a reference source when generating its output. The DRG prompt is designed to produce higher-quality outputs than the Baseline prompt, so the AI rater's ability to distinguish quality differences can be tested against a known contrast. Reference documentation curation, source materials, and clinical validation are provided in Appendix~\ref{app:refdoc}.

\subsection{Rater-Side Prompt Factors}
\label{sec:methods-rater-prompt}

The AI rater prompt is assembled from a modular block architecture with three rater-side factors: prompt character (lenient / moderate / strict, setting the level of detail required for a full score), prompt type (Normal / CoT / Self-Consistency, setting the reasoning architecture used before score commitment), and prompt order (three pre-specified orderings of the seven rater-prompt sections, holding section content fixed). Full prompt templates, the CoT audit categories, the SC safety-veto rule, and the prompt-order specifications are provided in Appendix~\ref{app:prompts}.

\subsection{Experimental Design and Analytical Framework}
\label{sec:methods-design}

The experimental design has three tiers. The patient tier consists of 16 synthetic patients. The CDSS tier crosses 4 generator models with 2 prompt configurations (DRG, Baseline), yielding 8 CDSS outputs per patient. The AI rater tier crosses 4 rater models, 3 prompt characters, 3 prompt types, and 3 prompt orders. To capture sampling variability under non-zero temperature, each fully specified design cell is executed for 3 independent runs at sampling temperature 0.7 (Appendix~\ref{app:models-inference}). The protocol factor (GR, Non-GR) is applied across all combinations, yielding $16 \times 8 \times 4 \times 3 \times 3 \times 3 \times 3 \times 2 = 82{,}944$ rating sessions, each conducted across all seven evaluation questions for $580{,}608$ question-level observations.

For each evaluation question, we fit a pooled linear mixed effects model (LMM) with the protocol factor and the five non-protocol factors as fixed effects, all two-way protocol $\times$ factor interactions, and three random-effect components (patient, prompt order, runs). Parameters are estimated by restricted maximum likelihood (REML) [23]. For each fixed-effect contrast we report the point estimate $\hat\tau_q$ with its 95\% Wald CI and Dunnett-adjusted $p$-value [24]; at the factor level we apply the Type 3 joint Wald test [25]. Per-level contrasts are summarized across the seven questions by inverse-variance weighting, and per-question $p$-values are combined by Fisher's method [26]. The dominant interaction (selected by largest joint Wald $\chi^2/\mathrm{df}$) is decomposed by per-level difference-in-differences (DID) [27] and per-level amplification ratio ($\mathrm{AR}_k = \hat\Delta^{\mathrm{GR}}_{k-\mathrm{ref}} / \hat\Delta^{\mathrm{Non\text{-}GR}}_{k-\mathrm{ref}}$), with AR reported as undefined when $|\hat\Delta^{\mathrm{Non\text{-}GR}}_{k-\mathrm{ref}}| < 0.5$ score units. Per-question Kolmogorov--Smirnov statistics [28] capture distributional differences beyond the mean shift. Full LMM specification, REML optimization, cross-question pooling formulae, multiplicity correction, DID/AR definitions, and the five per-rater behavioral dimensions are detailed in Appendix~\ref{app:lmm} (\S\S~\ref{app:lmm-spec}--\ref{app:lmm-mult}) and Appendix~\ref{app:rater-defs}.

%==================================================================
\section{Results \& Analysis}
\label{sec:results}

\subsection{Score Distribution under the Two Protocols}
\label{sec:results-dist}

\begin{figure}[!tbp]
\centering
\includegraphics[width=\linewidth]{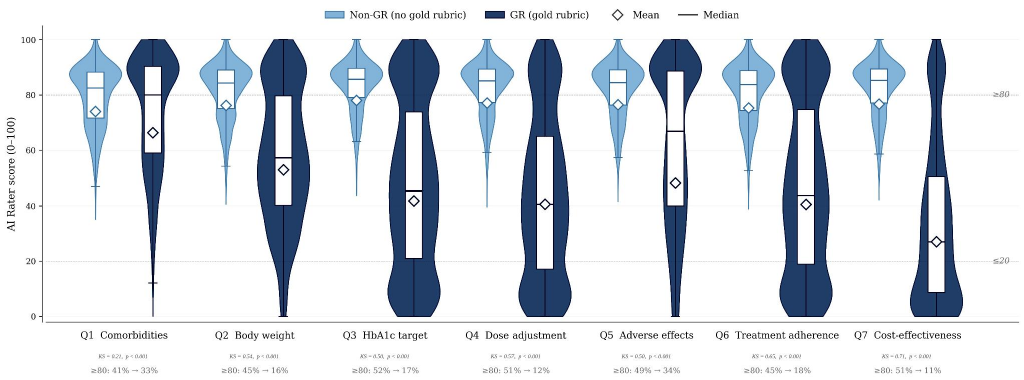}
\caption{Score distribution under Non-GR and GR protocols by evaluation question. Paired violin plot per question (Q1--Q7)---Non-GR (left, light blue) vs.\ GR (right, dark navy). White diamonds mark estimated mean scores, box overlays mark median and IQR, and dotted reference lines at 20 and 80 demarcate the floor and ceiling bands. Per-question Kolmogorov--Smirnov statistic, $p$-value, and change in ceiling-band ($\geq 80$) fraction are annotated below each question. GR scores are systematically lower and substantially more dispersed than Non-GR scores across all questions, with the protocol-induced shift varying approximately sevenfold (smallest at Q1, largest at Q7). Per-question numerical summaries: Appendix~\ref{app:numerical-dist}, Table~\ref{tab:score-dist}.}
\label{fig:dist}
\end{figure}

AI raters' scores had different distributions under the two protocols (all $p < 0.001$ from per-question KS tests; Figure~\ref{fig:dist}; Table~\ref{tab:score-dist}). Mean scores under GR (27.03--66.30) were lower than those under Non-GR (73.98--78.01) across all evaluation questions (all $p < 0.001$), but mean differences varied approximately sevenfold, with Q1 exhibiting the smallest shift ($\tau_{\text{protocol}} = -7.69$, 95\% CI [$-7.89$, $-7.50$]) and Q7 the largest ($\tau_{\text{protocol}} = -49.64$, 95\% CI [$-49.87$, $-49.42$]). We refer to this systematic difference between the two scoring protocols as the protocol-induced shift throughout the paper. The proportion of scores in the ceiling band ($\geq 80$) declined from 40.7--52.1\% under Non-GR to 11.4--33.9\% under GR across all questions, reflecting the consistently elevated score levels observed under Non-GR regardless of the evaluation question.

In addition, the interquartile range (IQR) of Non-GR scores fell consistently within the narrow band of 66--90 across all questions, whereas the GR condition yielded markedly heterogeneous IQRs that varied substantially by question. The size of IQR under GR was 1.68 to 3.67 times wider than that under Non-GR, indicating substantially greater score dispersion across all evaluation questions. The two protocols thus produced differences both in the magnitude of scores assigned by AI raters and in the degree of score dispersion across the seven evaluation questions---patterns that the interaction analyses in \S\ref{sec:results-cdss}--\S\ref{sec:results-rater} attribute to specific design factors.

\subsection{Fixed-Effect Component Analysis}
\label{sec:results-fixed}

\begin{figure}[!tbp]
\centering
\includegraphics[width=0.62\linewidth]{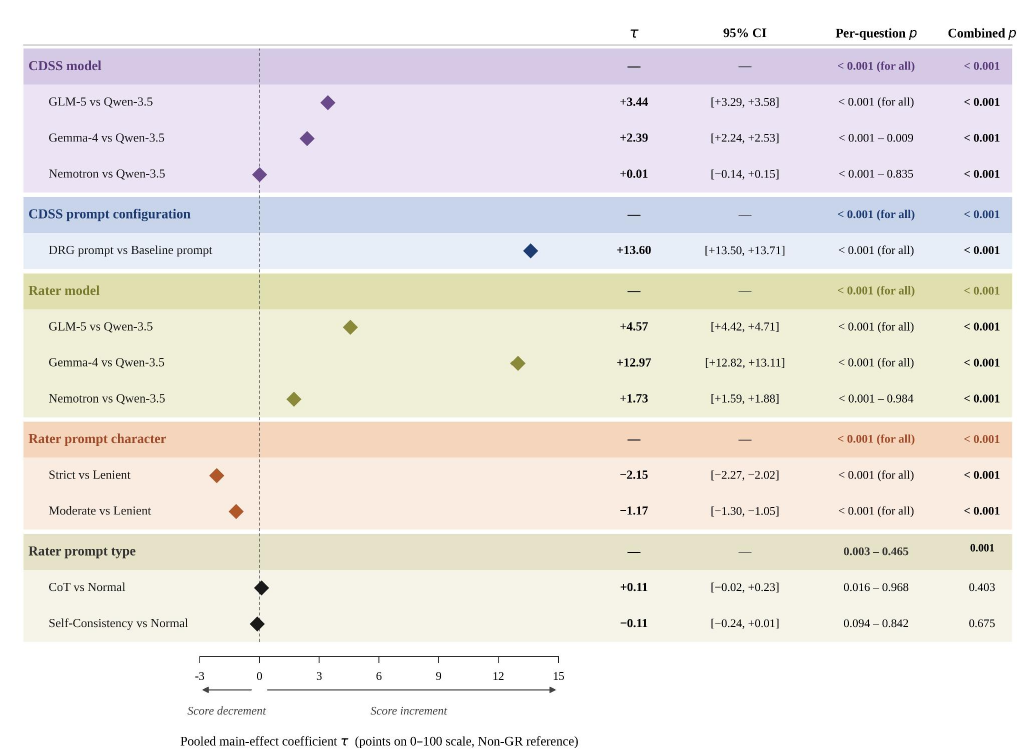}
\caption{Pooled main-effect estimates for the five non-protocol fixed-effect factors. Forest plot of $\tau$ coefficients showing each non-reference level relative to its reference (Baseline, Qwen-3.5, Lenient, Normal): diamonds mark point estimates; horizontal bars mark 95\% Wald CIs; the vertical dashed line marks $\tau = 0$. Bold rows at the top of each factor band report the factor-level Type 3 Wald joint test. CDSS prompt configuration (DRG vs.\ Baseline) and rater-model identity (Gemma-4 vs.\ Qwen-3.5) carry dominant main effects of comparable magnitude ($\sim 13$ score points), while prompt-type effects are roughly two orders of magnitude smaller. Question-level estimates and Dunnett-adjusted $p$-values: Appendix~\ref{app:numerical-pooled}, Table~\ref{tab:pooled-main}.}
\label{fig:forest}
\end{figure}

Among the five non-protocol factors, four reached factor-level significance on the precision-weighted pooled estimates (per-question $p < 0.001$ for all; Fisher-combined $p < 0.001$), while prompt type alone showed varied per-question significance ($p = 0.003$ to $0.465$; combined $p = 0.001$; Figure~\ref{fig:forest}). CDSS prompt configuration (DRG vs.\ Baseline) carried the largest main effect ($\hat\tau_{\text{cdss\_prompt}} = 13.60$, 95\% CI [13.50, 13.71]), with DRG outputs scored 13.60 points higher on average than Baseline. At the CDSS model level, outputs from GLM-5 ($\hat\tau = 3.44$, 95\% CI [3.29, 3.58]) and Gemma-4 ($\hat\tau = 2.39$, 95\% CI [2.24, 2.53]) received higher scores than those from the Qwen-3.5 reference, while Nemotron remained near the reference ($\hat\tau = 0.01$, 95\% CI [$-0.14$, 0.15]). Among rater models, AI raters built on Gemma-4 scored 12.97 points higher than those built on Qwen-3.5 ($\hat\tau = 12.97$, 95\% CI [12.82, 13.11]), with GLM-5 ($\hat\tau = 4.57$) and Nemotron ($\hat\tau = 1.73$) also higher than the Qwen-3.5 reference. Rater prompt character produced a monotonic decrease in scores as stringency increased, with Strict ($\hat\tau = -2.15$) and Moderate ($\hat\tau = -1.17$) both lower than the Lenient reference.

\subsection{Protocol $\times$ Factor Interactions: CDSS-Side}
\label{sec:results-cdss}

\begin{figure}[!tbp]
\centering
\includegraphics[width=0.7\linewidth]{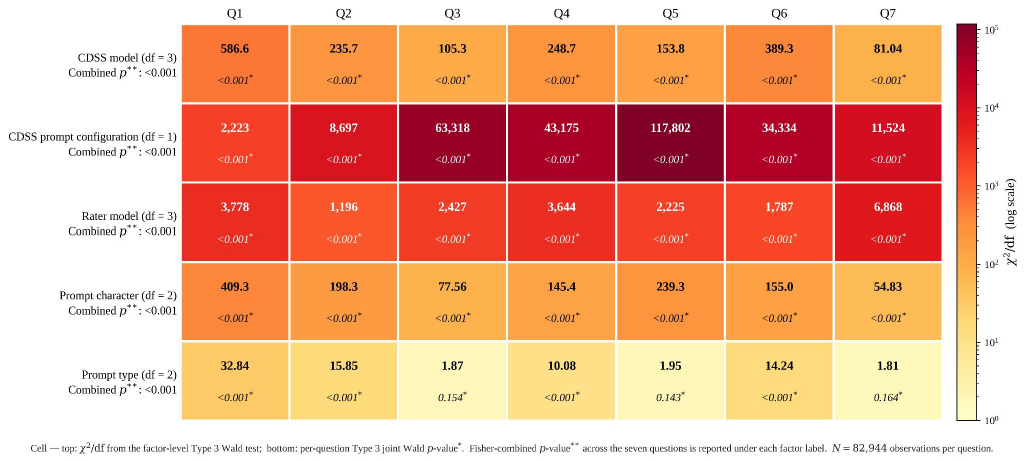}
\caption{Protocol $\times$ factor interaction heatmap across the seven evaluation questions. Each cell shows the per-question Type 3 Wald $\chi^2/\mathrm{df}$ statistic (top) and the per-question Type 3 joint Wald $p$-value (bottom); cell color encodes $\chi^2/\mathrm{df}$ on a log scale. Row labels include each factor's degrees of freedom and its Fisher-combined $p$-value across Q1--Q7. The scoring protocol moderates every non-protocol factor's effect; the protocol $\times$ CDSS prompt configuration interaction is the dominant interaction in every question, exceeding the others by an order of magnitude. $n = 82{,}944$ observations per question.}
\label{fig:heatmap}
\end{figure}

\begin{figure}[!tbp]
\centering
\includegraphics[width=0.85\linewidth]{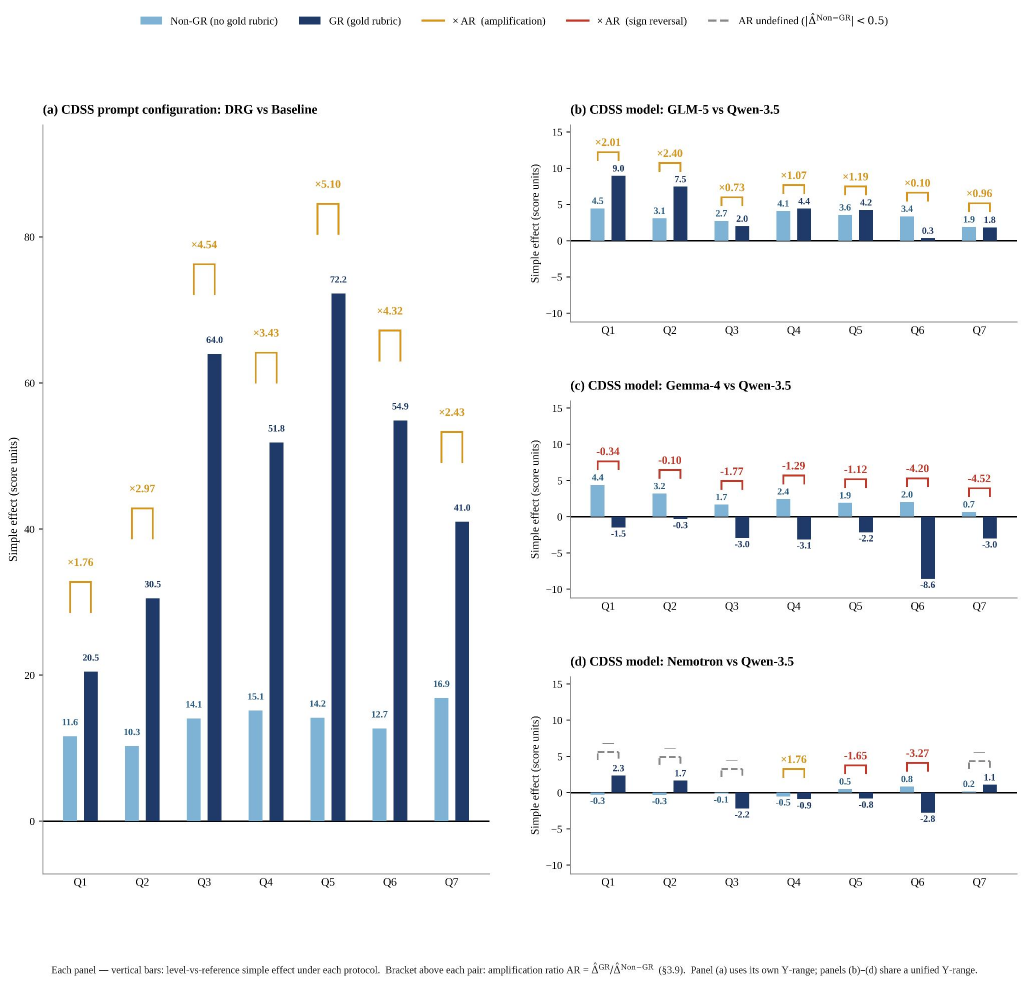}
\caption{Per-level decomposition of CDSS-side protocol $\times$ factor interactions. Four panels: (a) CDSS prompt configuration (DRG vs.\ Baseline); (b)--(d) CDSS model (GLM-5, Gemma-4, Nemotron each vs.\ Qwen-3.5). Each panel shows per-question simple effects (level vs.\ reference) under Non-GR (navy) and GR (maroon). The bracket above each bar pair reports the per-question amplification ratio (AR; Appendix~\ref{app:lmm-strat}), color-coded amber for amplification, red for sign reversal, and gray dashed when AR is undefined. Panel (a) uses its own Y-range; panels (b)--(d) share a unified Y-range. GR uniformly widens the DRG--Baseline gap across all questions (AR $\times 1.76$ to $\times 5.10$), but reshapes CDSS-model contrasts non-uniformly---including consistent sign reversal for Gemma-4 vs.\ Qwen-3.5 across every question. Per-level values: Appendix~\ref{app:numerical-decomp}.}
\label{fig:cdss-decomp}
\end{figure}

\textbf{Factor-level Type 3 Wald tests.} All five protocol $\times$ factor interactions reached Fisher-combined $p < 0.001$ across the seven evaluation questions, indicating that the scoring protocol moderated the effect of every non-protocol factor across questions (Figure~\ref{fig:heatmap}). Among the five, the protocol $\times$ CDSS prompt configuration interaction carried the largest joint Wald $\chi^2/\mathrm{df}$ in every question (2{,}223 to 117{,}802), exceeding every other interaction by an order of magnitude; the protocol $\times$ CDSS model interaction operated at smaller but still substantial magnitude ($\chi^2/\mathrm{df}$ = 81 to 587). To examine how the two scoring protocols reshape these effects, we decompose each interaction into per-level simple effects under Non-GR and GR (Figure~\ref{fig:cdss-decomp} for CDSS-side; Appendix~\ref{app:numerical-decomp}, Figure~\ref{fig:rater-decomp} for rater-side).

\textbf{The protocol $\times$ CDSS prompt configuration interaction.} Switching from Non-GR to GR widened the DRG--Baseline gap across all questions (Figure~\ref{fig:cdss-decomp}a). Simple effects of 10.27 to 16.86 score units under Non-GR expanded to 20.45 to 72.22 under GR, corresponding to ARs of $\times 1.76$ (Q1) to $\times 5.10$ (Q5). This pattern is attributable to the alignment between DRG outputs and rubric criteria: under GR, the rater has access to patient-specific rubric criteria, enabling it to acknowledge the additional rubric coverage inherent in DRG outputs and thereby widening the DRG--Baseline gap; under Non-GR, the absence of such criteria attenuates this differential, resulting in a substantially smaller scoring gap despite the same underlying difference in output quality.

\textbf{The protocol $\times$ CDSS model interaction.} Switching from Non-GR to GR did not act uniformly across CDSS models (Figure~\ref{fig:cdss-decomp}b--d). GLM-5 vs.\ Qwen-3.5 DID amplified at Q1 (AR = 2.01) and Q2 (AR = 2.40) but attenuated or reversed elsewhere (AR = 0.10 at Q6). Gemma-4 vs.\ Qwen-3.5 DID reversed sign across all questions (AR from $-4.52$ to $-0.10$), indicating that GR assigned lower scores to Gemma-4 outputs than to Qwen-3.5 outputs, whereas Non-GR scored Gemma-4 outputs higher than Qwen-3.5 outputs. Nemotron vs.\ Qwen-3.5 DID was numerically unstable in four of seven questions because the Non-GR contrast was already near zero. Unlike CDSS prompt configuration, the protocol effect for CDSS model is not uniform in direction across all models; rather, it depends on which CDSS model family is being evaluated.

\subsection{Protocol $\times$ Factor Interactions: Rater-Side}
\label{sec:results-rater}

\textbf{The protocol $\times$ rater model interaction.} Switching from Non-GR to GR produced three distinct patterns of rater-model contrast reshaping (Appendix~\ref{app:numerical-decomp}, Figure~\ref{fig:rater-decomp}a--c). The GLM-5 vs.\ Qwen-3.5 DID amplified consistently across all questions (AR = 1.57 to 6.58; largest DID = 16.86 at Q7). The Gemma-4 vs.\ Qwen-3.5 DID diverged into two regimes: DIDs at Q1 and Q2 attenuated nearly to zero (AR = 0.06 and 0.09), whereas Q3--Q7 showed sign reversal (AR = $-0.61$ to $-0.01$)---Non-GR simple effects (11.5 to 14.7) collapsed under GR ($-7.1$ to 1.1). The Nemotron vs.\ Qwen-3.5 DID amplified substantially at Q1--Q4 (AR up to 6.32) and Q7 (AR = $-17.23$), but was unstable at Q5 and Q6 owing to near-zero Non-GR contrasts. Rater model identity thus does not provide a uniform, transferable correction across protocols; rather, the direction of the protocol effect is jointly determined by rater model and evaluation question.

\textbf{The protocol $\times$ prompt character interaction.} Switching from Non-GR to GR uniformly amplified prompt character contrasts across all questions and levels (Appendix~\ref{app:numerical-decomp}, Figure~\ref{fig:rater-decomp}d--e), with ARs ranging from 2.32 to 3.62 and no sign reversal or unstable cells. A Strict instruction produced up to 3.6-fold stronger score shifts when the rubric was supplied, indicating that prompt-level calibration directives operate additively on top of the rubric rather than substituting for it: it issued without rubric context achieves only a fraction of its rubric-anchored effect.

\textbf{The protocol $\times$ prompt type interaction.} This was the smallest interaction in the design (Appendix~\ref{app:numerical-decomp}, Figure~\ref{fig:prompt-type}), with three questions failing to reach significance under per-question testing (Q3 $p = 0.154$; Q5 $p = 0.143$; Q7 $p = 0.164$). Per-level decomposition was uninformative across all contrasts, as the Non-GR simple effects were below 0.5 score units in every contrast, rendering ARs undefined. The protocol effect was not substantive for this factor; structured deliberation reorganizes how the rater applies its existing knowledge but does not supply patient-specific criteria.

\subsection{Per-Rater Behavioral Analyses}
\label{sec:results-perrater}

\textbf{Cross-rater consistency, mean shift, and reproducibility.} Protocol-induced shifts in mean score (Appendix~\ref{app:rater-mean}, Figure~\ref{fig:perrater}a) varied across the four raters---largest for Gemma-4 (45.72 points) and smallest for Nemotron (22.30 points; Table~\ref{tab:rater-shift})---and were consistent across CDSS models within each rater model rather than localized to specific rater--generator pairings (Figure~\ref{fig:cross-role}, Appendix~\ref{app:rater-cross}), so per-rater shifts generalize across CDSS sources. Run-to-run reproducibility was rater-specific and protocol-conditional: score variability remained stable across protocols for Gemma-4 ($\mathrm{SD}\ 3.17 \to 3.12$), whereas it doubled for Nemotron with protocol switching ($\mathrm{SD}\ 5.69 \to 12.64$; $p < 0.001$). For some raters the protocol therefore reshapes not only score level but also scoring stability---a property that would remain undetected in any single-protocol benchmark (Table~\ref{tab:rater-reprod}; Figure~\ref{fig:perrater}b).

\textbf{Responsiveness to rater-side prompt factors.} The Lenient--Strict score gap amplified under GR for every rater (Figure~\ref{fig:perrater}c; Non-GR shifts of 1.3--3.5 points expanded to 4.2--8.2 points under GR; Table~\ref{tab:rater-character}): a calibration directive (``be strict'') has no rubric-level anchor to operate on in its absence. In contrast, prompt-type effects remained an order of magnitude smaller across all rater $\times$ protocol cells (Figure~\ref{fig:perrater}d; within $\pm 2.7$ points overall, and within $\pm 1.0$ in seven of eight cells; Table~\ref{tab:rater-type}). Structured deliberation does not transfer rubric-equivalent calibration to the rater side.

\textbf{Self-preference under each protocol.} The self-preference effect was rater-specific and protocol-dependent (Table~\ref{tab:self-pref}; Figure~\ref{fig:perrater}e). Gemma-4 reversed sign across protocols ($+0.89 \to -4.38$; both $p < 0.001$), GLM-5 maintained a consistent positive effect on both protocols, while the remaining two raters were compatible with self-neutrality under at least one protocol. Self-preference---reported as a stable rater-side bias in general-domain LLM judges [8]---therefore does not generalize across the protocol axis in this clinical evaluation.

%==================================================================
\section{Limitations}
\label{sec:limitations}

Several limitations bound the scope of the present findings. First, this study reports quantitative comparisons of AI rater scores across scoring protocols but does not examine the internal reasoning processes underlying the observed patterns; understanding rater-side mechanisms is left to future work. Second, the experimental task is restricted to adult T2D pharmacotherapy at 12-month outpatient follow-up; how the protocol effect translates to clinical tasks with different decision structures remains an open question. Third, the four open-source LLMs in the rater pool exhibited substantial behavioral variation (\S\ref{sec:results-perrater}); caution is therefore warranted in interpreting these findings as broadly representative of LLM-as-a-judge behavior. Fourth, AI raters operate under fundamentally different scoring procedures depending on the protocol assigned (\S\ref{sec:methods-protocol}, Appendix~\ref{app:elicit-impl}); the protocol contrast reported here should accordingly be interpreted as a comparison of score distributions and directional shifts, rather than a comparison of absolute score values between the two protocols.

%==================================================================
\section{Conclusion}
\label{sec:conclusion}

The scoring protocol is not a downstream implementation choice in AI rater-based clinical evaluation; it fundamentally determines whether the evaluation can register meaningful differences among CDSS outputs. Rater-side prompt engineering, model selection, and structured deliberation strategies function as secondary controls rather than substitutes for the patient-specific rubric. The factorial design further demonstrates that AI rater's behavior is itself an object of study requiring systematic characterization, and should not be assumed to be stable across pipelines. Future work should disentangle the contribution of rubric content from that of the scoring procedure, validate rubric-anchored scoring against clinician judgment, and extend characterization to additional clinical domains and rater architectures, so that AI rater pipelines can be configured on the basis of documented behavioral evidence rather than assumed defaults.

%==================================================================
\section*{Acknowledgments and Disclosure of Funding}

\textbf{Funding.} This work was supported by GPU resources provided through the Advanced GPU Utilization Support Program, operated by the National IT Industry Promotion Agency (NIPA) and funded by the Ministry of Science and ICT (MSIT) of the Republic of Korea (Grant No.\ 02-26-01-0051). GPU infrastructure was provided by NHN Cloud. The funders had no role in study design, data collection and analysis, decision to publish, or preparation of the manuscript.

\textbf{Acknowledgments.} We thank the clinical staff at Samsung Medical Center for reviewing and validating the clinical appropriateness of the Type 2 Diabetes pharmacotherapy vignettes used in this study.

\textbf{Competing interests.} Sangwon Baek is the founder and chief executive officer of Asclep Korea Inc., a medical AI company. The AI rater evaluation framework examined in this work informs Asclep Korea's internal research and development; however, no commercial product is described, endorsed, or evaluated in this manuscript. Kyunga Kim and Kyu Yeon Hur declare no competing interests.

%==================================================================
\section*{References}
{\small

[1] Quinn, L., et al.\ (2023) Interobserver Variability Studies in Diagnostic Imaging: A Methodological Systematic Review. \textit{British Journal of Radiology}, 96(1148), 20220972.

[2] Di Forti, C.~L., et al.\ (2025) Inter-Rater Reliability of Psychiatric Diagnosis: A Systematic Review and Meta-Analysis. \textit{European Psychiatry}, 68(suppl.\ 1), S191--S192.

[3] Tawfik, D.~S., et al.\ (2018) Physician Burnout, Well-being, and Work Unit Safety Grades in Relationship to Reported Medical Errors. \textit{Mayo Clinic Proceedings}, 93(11), 1571--1580.

[4] Bedi, S., et al.\ (2026) Holistic Evaluation of Large Language Models for Medical Tasks with MedHELM. \textit{Nature Medicine}, 32(3), 943--951.

[5] Arora, R.~K., et al.\ (2025) HealthBench: Evaluating Large Language Models Towards Improved Human Health. arXiv:2505.08775.

[6] Kim, Y., et al.\ (2025) Medical Hallucination in Foundation Models and Their Impact on Healthcare. medRxiv:2025.02.28.25323115.

[7] Omar, M., et al.\ (2025) Multi-Model Assurance Analysis Showing Large Language Models Are Highly Vulnerable to Adversarial Hallucination Attacks During Clinical Decision Support. \textit{Communications Medicine}, 5, article 330.

[8] Panickssery, A., et al.\ (2024) LLM Evaluators Recognize and Favor Their Own Generations. \textit{Advances in Neural Information Processing Systems}, 37.

[9] Sharma, M., et al.\ (2024) Towards Understanding Sycophancy in Language Models. \textit{International Conference on Learning Representations}.

[10] Sclar, M., et al.\ (2024) Quantifying Language Models' Sensitivity to Spurious Features in Prompt Design or: How I Learned to Start Worrying about Prompt Formatting. \textit{International Conference on Learning Representations}.

[11] Zheng, L., et al.\ (2023) Judging LLM-as-a-Judge with MT-Bench and Chatbot Arena. \textit{Advances in Neural Information Processing Systems}, 36.

[12] Ye, J., et al.\ (2025) Justice or Prejudice? Quantifying Biases in LLM-as-a-Judge. \textit{International Conference on Learning Representations}.

[13] Koo, R., et al.\ (2024) Benchmarking Cognitive Biases in Large Language Models as Evaluators. \textit{Findings of the Association for Computational Linguistics: ACL 2024}, 517--545.

[14] Zhu, Z., et al.\ (2026) CyclicJudge: Mitigating Judge Bias Efficiently in LLM-based Evaluation. arXiv:2603.01865.

[15] Wang, P., et al.\ (2024) Large Language Models are not Fair Evaluators. \textit{Proceedings of the 62nd Annual Meeting of the Association for Computational Linguistics (Volume 1: Long Papers)}, 9440--9450.

[16] Li, Z., et al.\ (2024) Split and Merge: Aligning Position Biases in LLM-based Evaluators. \textit{Proceedings of the 2024 Conference on Empirical Methods in Natural Language Processing}, 11084--11108.

[17] Nasser, W.\ (2026) Evaluative Fingerprints: Stable and Systematic Differences in LLM Evaluator Behavior. arXiv:2601.05114.

[18] Greenblatt, R., et al.\ (2024) Alignment Faking in Large Language Models. arXiv:2412.14093.

[19] Serapio-Garc\'ia, G., et al.\ (2025) A Psychometric Framework for Evaluating and Shaping Personality Traits in Large Language Models. \textit{Nature Machine Intelligence}, 7, 1954--1968.

[20] Geathers, J., et al.\ (2025) Benchmarking Generative AI for Scoring Medical Student Interviews in Objective Structured Clinical Examinations (OSCEs). \textit{Artificial Intelligence in Education, Lecture Notes in Computer Science}, 15879, Springer, 231--245.

[21] Saito, K., et al.\ (2023) Verbosity Bias in Preference Labeling by Large Language Models. \textit{Workshop on Instruction Tuning and Instruction Following at NeurIPS 2023}.

[22] American Diabetes Association Professional Practice Committee. (2026) 9. Pharmacologic Approaches to Glycemic Treatment: Standards of Care in Diabetes---2026. \textit{Diabetes Care}, 49(suppl.\ 1), S183--S215. doi:10.2337/dc26-S009.

[23] Patterson, H.~D., \& Thompson, R.\ (1971) Recovery of Inter-Block Information When Block Sizes Are Unequal. \textit{Biometrika}, 58(3), 545--554.

[24] Dunnett, C.~W.\ (1955) A Multiple Comparison Procedure for Comparing Several Treatments with a Control. \textit{Journal of the American Statistical Association}, 50(272), 1096--1121.

[25] Pinheiro, J.~C., \& Bates, D.~M.\ (2000) \textit{Mixed-Effects Models in S and S-PLUS}. Springer.

[26] Fisher, R.~A.\ (1925) \textit{Statistical Methods for Research Workers}. Oliver and Boyd.

[27] Card, D., \& Krueger, A.~B.\ (1994) Minimum Wages and Employment: A Case Study of the Fast-Food Industry in New Jersey and Pennsylvania. \textit{American Economic Review}, 84(4), 772--793.

[28] Massey, F.~J.\ Jr.\ (1951) The Kolmogorov-Smirnov Test for Goodness of Fit. \textit{Journal of the American Statistical Association}, 46(253), 68--78.

[29] Ministry of Health and Welfare. (2026) Detailed Criteria and Methods for the Application of Health Insurance Benefits (Pharmaceutical): Partial Amendment. MOHW Notice No.~2026-24, Health Insurance Review \& Assessment Service, 29 Jan.\ 2026, \url{www.hira.or.kr}.

}

%==================================================================
\clearpage
\appendix

\section{Glossary of Terms}
\label{app:glossary}

\begin{table}[!htbp]
\centering
\caption{Glossary of study-specific terms.}
\label{tab:glossary}
\small
\begin{tabular}{p{0.30\linewidth} p{0.62\linewidth}}
\toprule
\textbf{Term} & \textbf{Definition} \\
\midrule
AI rater & LLM scoring CDSS outputs. \\
CDSS prompt configuration & The factor that controls how the CDSS generates its output; two levels (DRG, Baseline). \\
DRG & Document-referenced generation prompt; CDSS generates while using prepared reference documentation as a reference source. \\
Baseline prompt & CDSS generates from its parametric clinical knowledge without external documentation. \\
DRG--Baseline gap & Mean-score difference between DRG and Baseline outputs under a fixed scoring protocol. \\
Prompt character & Rater-side calibration directive (lenient / moderate / strict). \\
Prompt type & Rater-side reasoning strategy (Normal / CoT / Self-Consistency). \\
Prompt order & Random-effect factor for the sequence of rater-prompt sections. \\
GR protocol & Gold Rubric protocol---rater receives patient-specific rubric. \\
Non-GR protocol & Non Gold Rubric protocol---rater receives no rubric. \\
Pooled model & Single LMM per evaluation question with all design factors and scoring protocol $\times$ factor interactions. \\
Stratified model & LMM per evaluation question $\times$ CDSS prompt configuration stratum. \\
Protocol-induced shift & Systematic difference between scores produced under GR and Non-GR. \\
Amplification ratio (AR) & Ratio of the level-vs-reference simple effect under GR to the same effect under Non-GR. \\
Mean score & Average score (0--100 scale) that a rater assigns across all rating sessions in a given condition. \\
Run-to-run reproducibility & Within-condition standard deviation of a rater's scores across the three repeated runs at temperature 0.7. \\
Responsiveness to prompt character & Change in a rater's mean score when prompt character is shifted from Lenient to Strict. \\
Responsiveness to prompt type & Change in a rater's mean score when prompt type is shifted from Normal to either CoT or SC. \\
Self-preference effect & Difference between a rater's mean score on its own CDSS outputs and on the other three CDSS models' outputs. \\
\bottomrule
\end{tabular}
\end{table}

%==================================================================
\section{Score Elicitation Schemes}
\label{app:elicit}

\subsection{GR Elicitation: Rubric-Element Aggregation}
\label{app:elicit-gr}

Under GR, each evaluation question $q$ is decomposed into a set of rubric elements $e$ ($1 \leq e \leq E_q$) defined in the patient-specific rubric $R$. For each element, the AI rater assigns a discrete score $s_{q,e} \in \{0, 1, 2\}$, where 0 indicates absent or contradicted, 1 indicates partially addressed, and 2 indicates fully addressed. The element-level scores are aggregated as the sum of element scores divided by the maximum achievable sum ($2 \times E_q$), producing a unit-interval score $r_q \in [0, 1]$, then truncated to two decimals and rescaled by 100 to yield the final 0--100 scale value. The rater's degrees of freedom under GR are restricted to the discrete element-level decision.

\subsection{Non-GR Elicitation: Direct 0--100 Scoring}
\label{app:elicit-nongr}

Under Non-GR, the AI rater outputs an integer score $s_q \in \{0, 1, \dots, 100\}$ directly per evaluation question, without intermediate element-level decomposition or aggregation. The rater is provided with a verbal scoring rubric describing the qualitative meaning of representative score bands but is not asked to commit to discrete element-level decisions. The output is parsed as a single integer in the $[0, 100]$ range.

\subsection{Implications for Interpretation}
\label{app:elicit-impl}

The two elicitation schemes differ in three respects. First, GR aggregates discrete element-level decisions while Non-GR produces a single integer output. Second, GR scores fall on a discrete grid determined by the number of rubric elements; Non-GR scores cover all integers in $[0, 100]$. Third, the patient-specific scoring criteria are exposed to the rater under GR but remain implicit under Non-GR. These differences cannot be eliminated without altering the protocol manipulation itself: rubric content is operationalized as element-level criteria. The protocol effect quantified in \S\ref{sec:results} therefore reflects the joint effect of rubric content and elicitation format.

%==================================================================
\section{Patient Construction Protocol and Case Selection}
\label{app:cases}

\subsection{Evaluation Question Definitions}
\label{app:cases-questions}

The seven evaluation questions used throughout the study were defined through a modified Delphi procedure with three senior board-certified diabetologists (\S\ref{app:cases-consult}). Each question names a clinical decision dimension addressed by a prescribing diabetologist during a 12-month T2D outpatient follow-up visit, together with the operational scope under which the AI rater assesses the CDSS output. The full construction process---the axes of clinical variability used to instantiate evaluation cases, the consultation procedure, and the case-selection logic---is described in \S\S\ref{app:cases-axes}--\ref{app:cases-selection}.
\begin{itemize}
    \item \textbf{Q1: Comorbidities.} Whether the recommended pharmacotherapy reflects the patient's comorbidity profile (ASCVD, HF, MASLD, CKD).
    \item \textbf{Q2: Body weight.} Whether agent selection accounts for the patient's BMI category (weight-favorable, weight-neutral, or weight-promoting effect).
    \item \textbf{Q3: HbA1c target.} Whether the HbA1c target is individualised against the patient's age, comorbidity burden, hypoglycemia risk, and life expectancy.
    \item \textbf{Q4: Dose adjustment.} Whether the recommendation specifies starting dose, titration schedule, and renal-adjustment cut-offs where applicable.
    \item \textbf{Q5: Adverse effects.} Whether known adverse-effect risks for the chosen agents are flagged and contraindications are explicitly excluded.
    \item \textbf{Q6: Treatment adherence.} Whether the regimen accounts for adherence barriers (dosing frequency, injection burden, polypharmacy).
    \item \textbf{Q7: Cost-effectiveness.} Whether the choice satisfies Korean Health Insurance Review \& Assessment Service (HIRA) reimbursement criteria [29] for the patient's clinical profile.
\end{itemize}

\subsection{Axes of Clinical Variability}
\label{app:cases-axes}

The sixteen synthetic patients were constructed to span the clinical variability encountered in T2D outpatient pharmacotherapy decision making. The senior board-certified diabetologists defined three categories of axes and their levels.

Comorbidity profile comprises three binary axes---atherosclerotic cardiovascular disease (ASCVD), heart failure (HF), and metabolic dysfunction-associated steatotic liver disease (MASLD)---each coded as present or absent.

Metabolic status is captured by HbA1c level, stratified into four clinically meaningful bands: 5.5--6.5\% (near-target), 6.5--7.5\% (mild hyperglycemia), 7.5--9.0\% (moderate hyperglycemia), and 9.0--11.0\% (severe hyperglycemia).

Anthropometric category is captured by body-mass index (BMI), stratified into three bands at clinically relevant thresholds: 18--23~kg/m\textsuperscript{2} (normal range), 23--27~kg/m\textsuperscript{2} (overweight), and 27--35~kg/m\textsuperscript{2} (obesity range).

Each axis modulates evidence-based pharmacotherapy choices in T2D: ASCVD and HF status shape SGLT2 inhibitor and GLP-1 receptor agonist indications; MASLD status affects metformin and pioglitazone considerations; HbA1c band determines treatment intensity; and BMI category informs agent selection around body-weight-favorable therapies.

\subsection{Iterative Expert Consultation}
\label{app:cases-consult}

Axis definitions and level cut-offs were finalized through a modified Delphi procedure with three senior board-certified diabetologists. Each round generated candidate axis definitions and cut-off values, which were reviewed for clinical plausibility, discriminative relevance, and alignment with current practice guidelines. Iteration continued until consensus.

\subsection{Case Selection Logic}
\label{app:cases-selection}

The full factorial combination of the axes above produces a candidate pool of 480 cases. From this pool, sixteen cases were selected to serve as the evaluation patient set under two principles. Representativeness requires that selected cases reflect clinical presentations encountered in routine T2D outpatient care rather than artificially constructed edge combinations. Coverage of the decision surface requires that the selected cases jointly exercise the clinically meaningful regions of the axis space, so that Q1--Q7 are evaluated across a range of pharmacotherapy decision contexts.

The reduction from 480 candidates to 16 patients also reflects the computational footprint of the factorial design. Running the full design (8 CDSS outputs $\times$ 4 raters $\times$ 3 prompt characters $\times$ 3 prompt types $\times$ 3 prompt orders $\times$ 3 runs $\times$ 2 scoring protocols = 5{,}184 rating sessions per patient) across all 480 candidates would require 2{,}488{,}320 rating sessions, infeasible at the available compute scale.

%==================================================================
\section{Model Architecture and Inference Configuration}
\label{app:models}

\subsection{Model Architectures}
\label{app:models-arch}

The four open-source LLMs serve as both CDSS models and AI raters under the dual-role design. The four models were selected to span architectural diversity on three axes. Parameter scale ranges from 30.7~billion (Gemma-4) to 744~billion (GLM-5), covering more than an order of magnitude. The attention mechanism varies from standard dense attention (Gemma-4) through grouped-query attention (Qwen-3.5) to sparse-attention variants (GLM-5 with Deep Sparse Attention; Nemotron-3 with hybrid Transformer--Mamba layers and Multi-Token Prediction). The sparsity structure ranges from fully dense (Gemma-4) to mixture-of-experts with varying activation ratios (approximately 4.2\% active in Qwen-3.5 to 9.7\% in Nemotron-3). This diversity allows factor-level inferences about AI rater behavior to extend beyond a single architectural family.

Each model is publicly released under the following license; the Hugging Face Hub identifiers below refer to the NVFP4 quantized variants distributed by NVIDIA and used in deployment (see \S\ref{app:models-inference} on inference precision). Full architectural specifications (layer count, hidden size, attention configuration, expert count and activation ratio for MoE variants, training-token budget) are provided in the project release.
\begin{itemize}
    \item \textbf{Qwen-3.5} --- Apache License 2.0; \path{nvidia/Qwen-3.5-397B-A17B-NVFP4} on Hugging Face Hub.
    \item \textbf{GLM-5} --- MIT License; \path{nvidia/GLM-5-NVFP4} on Hugging Face Hub.
    \item \textbf{Gemma-4} --- Apache License 2.0; \path{nvidia/Gemma-4-31B-IT-NVFP4} on Hugging Face Hub.
    \item \textbf{Nemotron-3-Super-120B-A12B} --- NVIDIA Nemotron Open Model License; \path{nvidia/NVIDIA-Nemotron-3-Super-120B-A12B-NVFP4} on Hugging Face Hub.
\end{itemize}

\subsection{Inference Configuration}
\label{app:models-inference}

All four models were deployed under matched inference settings, so that observed behavioral differences reflect architectural rather than configuration variation. Sampling temperature was held at 0.7 across all generation and scoring runs; top-$p$ was set to 0.95; maximum generation length was 128{,}000 tokens for CDSS outputs and 65{,}536 tokens for AI rater outputs. Each condition was run three times per cell with independent random seeds to capture sampling-induced variation. All models were deployed at FP4 precision via NVIDIA NVFP4 quantized variants on a tensor-parallel cluster with sixteen NVIDIA B200 GPUs per model instance, distributed via the Hugging Face Hub.

%==================================================================
\section{Reference Documentation Curation for Document-Referenced Generation}
\label{app:refdoc}

\subsection{Source Materials and Scope}
\label{app:refdoc-source}

The reference documentation supplied to the CDSS under the DRG prompt configuration was extracted from two authoritative source materials. The first is the American Diabetes Association \emph{Standards of Care in Diabetes---2026}, Section~9 (Pharmacologic Approaches to Glycemic Treatment) [22]. Section~9 was the only section used; sections covering screening, monitoring, lifestyle management, and other non-pharmacotherapy topics were excluded as outside the evaluation task scope. The second is the Korean HIRA reimbursement framework, MOHW Notice No.~2026-24 [29], used specifically for the cost-effectiveness question (Q7). No other source materials contributed to the reference documentation.

\subsection{Multi-Stage Extraction Procedure}
\label{app:refdoc-extract}

Extraction was conducted using Claude Opus~4.7 through a multi-stage procedure. Each clinical decision area covered by the seven evaluation questions was processed independently, producing a self-contained reference document per area. The first pass identified relevant guideline passages and reimbursement criteria; subsequent passes refined the extracted content for completeness, removed redundancy, and verified that every recommendation traced back to a specific source-material location. The extraction prompts and intermediate outputs are not released; the reference documents are described structurally in \S\ref{app:refdoc-structure} and may be requested from the corresponding author.

\subsection{Clinical Validation}
\label{app:refdoc-validation}

All extracted reference documents were reviewed and validated by three senior board-certified diabetologists prior to use. Each reviewer confirmed traceability of every recommendation to its source-material location, flagged any recommendation whose extracted form altered the clinical meaning, and confirmed that no clinically essential content was omitted. Flagged recommendations were revised and re-reviewed until all three reviewers reached agreement. The same diabetologists also validated the patient-specific rubric contents used under the GR protocol (\S\ref{sec:methods-protocol}), so the CDSS-side documentation and the rater-side rubric passed the same validation step.

\subsection{Structure of the Reference Documents}
\label{app:refdoc-structure}

Each reference document is organized in the order a prescribing clinician would address each decision during a follow-up visit, mirroring the seven evaluation questions. The document covers comorbidity-aligned drug class selection (Q1), body-weight-aware agent choice (Q2), patient-individualized glycemic targets (Q3), dose specification and titration (Q4), adverse-effect mitigation and contraindications (Q5), adherence-supportive regimen design (Q6), and reimbursement-aligned agent selection under HIRA criteria (Q7). The CDSS model receives the documentation alongside the patient case and uses it as a reference source when generating its treatment recommendation. The Baseline prompt configuration omits the documentation entirely; the model receives only the patient case and the generation instruction. Decision elements are aligned across the GR rubric and the DRG documentation so that the rubric and the documentation refer to the same underlying clinical criteria; the AI rater alone receives the rubric, and the CDSS alone receives the documentation.

%==================================================================
\section{Prompt Templates and Rater-Side Configuration}
\label{app:prompts}

\subsection{Modular Block Architecture}
\label{app:prompts-modular}

The AI rater prompt is composed of seven modular sections: role and scope, type-specific reasoning method, input contract, output structure, scoring rules, prohibitions, and quality assurance. The content of each section is held identical across all conditions except where explicitly varied by prompt character, prompt type, or prompt order.

The prompt character factor varies the content of the scoring rules and quality assurance sections only---specifically, the operational definition of what qualifies for a full score. Strict requires three explicit elements (a patient-specific anchor, a clinically appropriate action, and an actionable next step). Moderate requires one explicit anchor plus one actionable element. Lenient accepts clinically appropriate and safe recommendations even when anchors are implicit. The zero-score boundary---absent or unsafe recommendations---is held identical across the three levels, so that prompt character varies the rater's evaluative disposition only at the upper end of the scale.

\subsection{CoT Five-Stage Clinical Audit}
\label{app:prompts-cot}

Under the CoT prompt type, the rater completes a structured sequential audit before committing to a score. Given a patient $p$, a CDSS output $o$, an evaluation question $q$, and a rubric $R_q$, the rater performs five sub-audits: glycemic control ($a_1$), regimen appropriateness ($a_2$), renal safety ($a_3$), cardiorenal-weight considerations ($a_4$), and actionability ($a_5$). The rater aggregates the sub-audits into $\bar{a}$ and commits to a score $s = f(\bar{a}, R_q)$, with $R_q$ omitted under Non-GR. The audit trace $\bar{a}$ is suppressed from output to prevent length from confounding the score.

\subsection{SC Safety-Veto Aggregation}
\label{app:prompts-sc}

Under the SC prompt type, three internal judges independently score the output, after which their outputs are aggregated under a safety-veto rule. The three judges score anchor sufficiency ($j_1$), therapeutic fit ($j_2$), and safety ($j_3$) using the same inputs. The aggregation follows:
\begin{equation}
s_{\text{final}} \;=\;
\begin{cases}
j_3 & \text{if } j_3 \leq \tau, \\[4pt]
\dfrac{1}{3}(j_1 + j_2 + j_3) & \text{otherwise.}
\end{cases}
\label{eq:sc}
\end{equation}
The veto rule asymmetrizes the aggregation: only the safety judge can force the minimum score regardless of the other judges' votes; the other two judges contribute through averaging when the safety judge does not veto.

\subsection{Prompt Order}
\label{app:prompts-order}

The prompt-order factor was implemented as a controlled variation on section salience rather than an exhaustive permutation experiment. Within each prompt type and prompt character, the section inventory was held constant and only the order of the variable sections was changed. Three pre-specified orders were used, applied identically to both protocols.

\textbf{Order A (canonical task-flow):} role and scope, type-specific reasoning method, input contract, output structure, scoring rules, prohibitions, quality assurance.

\textbf{Order B (structure-first):} output structure, prohibitions, role and scope, type-specific reasoning method, input contract, scoring rules, quality assurance---output constraints and prohibitions are placed in the primacy position.

\textbf{Order C (scoring-first-after-role):} role and scope, scoring rules, input contract, type-specific reasoning method, output structure, prohibitions, quality assurance---scoring criteria are placed before the procedural method.

The role-and-scope and quality-assurance blocks are held in stable positions across all three orders to preserve task identity. Under Non-GR, the section inventory expands to include patient-first stance, scoring reference, direct scoring rules, and safety-override triggers, but the same A/B/C salience logic applies.

%==================================================================
\section{Linear Mixed Effects Model Specification (Full Detail)}
\label{app:lmm}

This appendix provides the full mathematical and procedural specification of the linear mixed effects modeling framework summarized in \S\ref{sec:methods-design} of the main text. Sections~\ref{app:lmm-spec}--\ref{app:lmm-mult} expand the pooled-model analysis, and Section~\ref{app:lmm-strat} expands the stratified analysis of the dominant interaction. The supporting-analyses paragraph is also provided in Section~\ref{app:lmm-support}.

\subsection{Pooled Model Specification}
\label{app:lmm-spec}

\textbf{Pooled model specification.} For each evaluation question, we fit a pooled LMM with the protocol factor and the five non-protocol factors as fixed effects, together with all two-way protocol $\times$ factor interactions. The pooled model decomposes the variation in AI rater scores into fixed-effect factors, random-effect components, and a residual:
\begin{equation}
\begin{aligned}
Y \;=\;& \mu + \tau_{\text{protocol}} + \tau_{\text{cdss\_model}} + \tau_{\text{cdss\_prompt}} + \tau_{\text{rater\_model}} \\
&{}+ \tau_{\text{prompt\_char}} + \tau_{\text{prompt\_type}} + \tau_{\text{interactions}} \\
&{}+ u_{\text{patient}} + u_{\text{order}} + u_{\text{runs}} + \varepsilon
\end{aligned}
\label{eq:lmm}
\end{equation}
$Y$ is the AI rater score on the 0--100 scale from a single rating session, and $\mu$ is the overall mean. Fixed-effect factors capture systematic sources of variation: $\tau_{\text{protocol}}$ denotes the protocol effect (GR vs.\ Non-GR); $\tau_{\text{cdss\_model}}$ the CDSS model effect; $\tau_{\text{cdss\_prompt}}$ the CDSS prompt configuration effect (DRG vs.\ Baseline); $\tau_{\text{rater\_model}}$ the rater model effect; $\tau_{\text{prompt\_char}}$ the prompt character effect; $\tau_{\text{prompt\_type}}$ the prompt type effect. $\tau_{\text{interactions}}$ collects the five two-way interactions between the protocol factor and each non-protocol factor. All fixed-effect terms are encoded using treatment coding, with reference levels specified at the end of this subsection. Three random-effect components are specified to account for variance attributable to distinct clustering units: $u_{\text{patient}}$ for patients, $u_{\text{order}}$ for prompt orders, and $u_{\text{runs}}$ for repeated runs, the last of which captures run-to-run scoring variability under sampling temperature 0.7. The residual $\varepsilon$ represents the remaining unexplained variation. All random-effect components and the residual are assumed to be mutually independent and normally distributed:
\begin{equation}
u_{\text{patient}} \sim \mathcal{N}(0, \sigma^2_{\text{patient}}), \quad
u_{\text{order}} \sim \mathcal{N}(0, \sigma^2_{\text{order}}), \quad
u_{\text{runs}} \sim \mathcal{N}(0, \sigma^2_{\text{runs}}), \quad
\varepsilon \sim \mathcal{N}(0, \sigma^2_{\varepsilon}).
\label{eq:re-distributions}
\end{equation}
Parameters are estimated by restricted maximum likelihood (REML) [23]; optimization details are provided in \S\ref{app:lmm-opt}. The pooled model is fit independently for each of the seven evaluation questions, yielding seven sets of fixed-effect estimates and random-effect variance components.

\textbf{Treatment-coding reference levels.} Scoring protocol = Non-GR; CDSS model = Qwen-3.5; CDSS prompt configuration = Baseline; Rater model = Qwen-3.5; Prompt character = Lenient; Prompt type = Normal; Prompt order = Order A.

\subsection{Estimation and Optimizer}
\label{app:lmm-opt}

\begin{sloppypar}
Random-effect components are estimated by restricted maximum likelihood (REML) using \texttt{statsmodels.MixedLM}, with a maximum of 200 iterations and a convergence tolerance of $10^{-6}$. All LMMs reached convergence on the first attempt without modification.
\end{sloppypar}

\textbf{Per-question estimates.} For each fixed-effect contrast (each non-reference level vs.\ its reference), we report the point estimate $\hat\tau_q$ with its 95\% Wald CI $\hat\tau_q \pm 1.96 \cdot \mathrm{SE}(\hat\tau_q)$ and the corresponding Dunnett-adjusted $p$-value [24] (\S\ref{app:lmm-mult}). At the factor level, we apply the Type 3 joint Wald test [25] to evaluate all $K-1$ non-reference contrasts simultaneously.

\subsection{Cross-Question Pooling}
\label{app:lmm-pool}

\textbf{Combining estimates across questions.} Per-level contrasts are summarized across the seven questions by inverse-variance weighting of $\hat\tau_q$, yielding a pooled estimate with 95\% Wald CI:
\begin{equation}
\hat\tau_{\mathrm{pool}} \;=\; \frac{\sum_{q=1}^{7} w_q \, \hat\tau_q}{\sum_{q=1}^{7} w_q}, \qquad w_q = \frac{1}{\mathrm{SE}^2(\hat\tau_q)}.
\label{eq:pool}
\end{equation}
The pooled standard error is $\mathrm{SE}(\hat\tau_{\mathrm{pool}}) = (\sum_q w_q)^{-1/2}$ and the pooled 95\% Wald CI is $\hat\tau_{\mathrm{pool}} \pm 1.96 \cdot \mathrm{SE}(\hat\tau_{\mathrm{pool}})$. Inverse-variance weighting yields the minimum-variance unbiased linear combination of the per-question estimates under the assumption that the seven $\hat\tau_q$ are independent. The seven per-question Dunnett-adjusted $p$-values are combined by Fisher's method [26] (\S\ref{app:lmm-mult}). At the factor level, the seven per-question Type 3 Wald $p$-values---each already a single composite test, with no Dunnett step in its derivation---are combined by the same Fisher procedure.

\subsection{Multiplicity Correction (Dunnett $+$ Fisher)}
\label{app:lmm-mult}

Multiplicity correction proceeds on two axes---within questions and across questions.

\textbf{Within questions.} Within each (question, factor) family, the $m$ level-vs-reference Wald tests share a common reference, which induces negative covariance among the contrasts. Dunnett's correction is applied within each family using the multivariate normal max-$|Z|$ distribution derived from the LMM contrast covariance, converting each per-question Wald $p$-value $p_{q,j}$ into a Dunnett-adjusted $p$-value $p_{q,j}^{\mathrm{Dunnett}}$. For binary factors ($m = 1$), Dunnett's adjustment reduces to the unadjusted Wald $p$-value.

\textbf{Across questions.} When a single cross-question summary $p$-value is reported, the seven per-question Dunnett-adjusted $p$-values for the same contrast are combined using Fisher's method:
\begin{equation}
X^2 \;=\; -2 \sum_{q=1}^{7} \ln\!\bigl(p_{q,j}^{\mathrm{Dunnett}}\bigr) \;\sim\; \chi^2_{14} \quad \text{under } H_0.
\label{eq:fisher}
\end{equation}
The same Fisher procedure is applied at the factor level to combine the seven per-question Type 3 Wald $p$-values into a single cross-question $p$-value summarizing each factor-level effect.

\subsection{Stratified Analysis of the Dominant Interaction}
\label{app:lmm-strat}

\textbf{Stratified model specification.} Among the protocol $\times$ factor interactions evaluated by the Fisher-combined Type 3 Wald $p$-values from \S\ref{app:lmm-spec}--\S\ref{app:lmm-mult}, the interaction with the largest joint Wald $\chi^2/\mathrm{df}$ is selected for follow-up analysis. We refit the pooled LMM separately within each level of that factor, omitting that factor and its protocol interaction from the fixed-effect specification while retaining the same random-effect structure. One model is fit per (question, stratum) combination.

\textbf{Per-level difference-in-differences (DID).} When a factor has more than two levels, the scoring protocol effect can vary across the levels of that factor. To quantify this variation, we report the per-level DID contrast [27]. For each non-reference level $k$ of any factor, let $\hat\Delta^{\mathrm{GR}}_{k-\mathrm{ref}}$ denote the estimated mean-score difference between level $k$ and the reference level under the GR protocol, and $\hat\Delta^{\mathrm{Non\text{-}GR}}_{k-\mathrm{ref}}$ denote the same difference under the Non-GR protocol. The DID contrast is
\begin{equation}
\widehat{\mathrm{DID}}_k \;=\; \hat\Delta^{\mathrm{GR}}_{k-\mathrm{ref}} - \hat\Delta^{\mathrm{Non\text{-}GR}}_{k-\mathrm{ref}}.
\label{eq:did}
\end{equation}
Each DID is reported with its Wald 95\% confidence interval and Dunnett-adjusted $p$-value. For a binary factor, the per-level DID reduces to a single contrast.

\textbf{Per-level amplification ratio (AR).} To express the protocol $\times$ factor interaction in a scale-free form, we additionally report the per-level AR:
\begin{equation}
\mathrm{AR}_k \;=\; \frac{\hat\Delta^{\mathrm{GR}}_{k-\mathrm{ref}}}{\hat\Delta^{\mathrm{Non\text{-}GR}}_{k-\mathrm{ref}}},
\label{eq:ar}
\end{equation}
the ratio of the level-vs-reference simple effect under GR to the same simple effect under Non-GR. $\mathrm{AR} > 1$ indicates that GR amplifies the level-vs-reference difference relative to Non-GR; $0 < \mathrm{AR} < 1$ indicates that GR attenuates it; $\mathrm{AR} < 0$ indicates a sign reversal between protocols. AR is reported as undefined when $|\hat\Delta^{\mathrm{Non\text{-}GR}}_{k-\mathrm{ref}}| < 0.5$ score units, since the ratio becomes numerically unstable in this regime. For a binary factor, AR reduces to the GR-side gap divided by the Non-GR-side gap.

\subsection{Supporting Analyses}
\label{app:lmm-support}

Two supporting analyses complement the main inference. Per-question Kolmogorov--Smirnov (KS) statistics [28] capture distributional differences in AI rater scores between protocols beyond the mean shift. Each rater's scoring behavior is characterized along five dimensions---mean score, run-to-run reproducibility, responsiveness to prompt character, responsiveness to prompt type, and self-preference effect (defined in Appendix~\ref{app:glossary}; full numerical records in Appendix~\ref{app:rater-defs}).

%==================================================================
\section{Detailed Numerical Results (Main Effects, Random Effects, and Factor-Level Interactions)}
\label{app:numerical-main}

\subsection{Score Distribution Under the Two Protocols}
\label{app:numerical-dist}

\begin{table}[!tbp]
\centering
\caption{Distribution of AI rater scores under the two scoring protocols by evaluation question. Columns report mean (SD) and median (IQR) of AI rater scores under each protocol, the Mean Difference between the two protocols (GR $-$ Non-GR) with its 95\% confidence interval, and the $p$-value for the difference. $n = 82{,}944$ observations per question.}
\label{tab:score-dist}
\includegraphics[width=\linewidth]{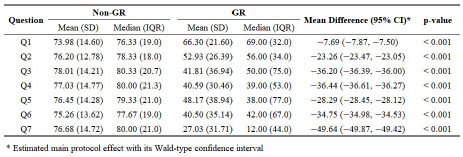}
\end{table}

Table~\ref{tab:score-dist} reports the per-question score distributions visualized in body Figure~\ref{fig:dist}. Mean scores under GR (27.03--66.30) are uniformly lower than those under Non-GR (73.98--78.01) across all seven questions, with all per-question Mean Differences reaching $p < 0.001$. The magnitude of the protocol-induced shift varies approximately sevenfold across questions---Q1 carries the smallest shift and Q7 the largest---establishing that the scoring protocol effect is question-dependent rather than a uniform additive offset. Per-question median and IQR values confirm that the protocol effect operates not only on the mean but on the upper-tail concentration of scores (Non-GR distributions are right-shifted with substantial mass above 80; GR distributions occupy a wider operational range across the 0--100 scale).

\subsection{Pooled Main Effects of Non-Protocol Factors}
\label{app:numerical-pooled}

\begin{table}[!tbp]
\centering
\caption{Pooled main-effect estimates for the five non-protocol fixed-effect factors. For each non-reference level relative to its reference, we report the pooled main effect ($\tau$) with 95\% Wald CI, the Fisher's combined $p$-value, and the range of per-question Dunnett-adjusted $p$-values. Reference levels: Baseline (CDSS prompt configuration), Qwen-3.5 (rater model and CDSS model), Lenient (prompt character), Normal (prompt type). $n = 82{,}944$ observations per question.}
\label{tab:pooled-main}
\includegraphics[width=\linewidth]{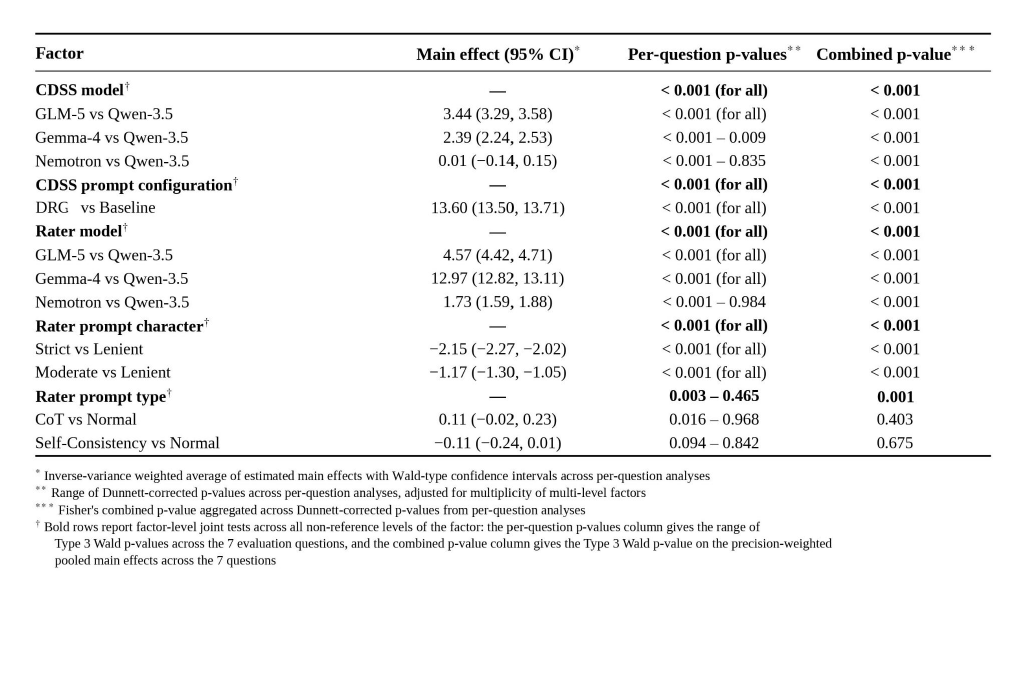}
\end{table}

Table~\ref{tab:pooled-main} reports $\tau$ pooled across the seven questions by inverse-variance weighted average, by factor and level, alongside factor-level joint Type 3 Wald tests on the pooled estimates. Two findings follow directly. First, the CDSS prompt configuration contrast (DRG vs.\ Baseline, $\tau = +13.60$) and the strongest rater-model contrast (Gemma-4 vs.\ Qwen-3.5, $\tau = +12.97$) operate at almost the same magnitude. Switching the rater model from Qwen-3.5 to Gemma-4 therefore shifts the AI rater's score by roughly the same amount as switching what the CDSS sees at generation time---placing rater-model identity on the same footing as CDSS prompt configuration. Second, the prompt-type factor carries the smallest joint effect among the five non-protocol factors (Wald $\chi^2 = 35.78$, df $= 14$, $p = 0.001$), with both per-level contrasts within $\pm 0.15$ of zero; the joint Wald $\chi^2$ is approximately two orders of magnitude smaller than those of the other four factors. Structured reasoning strategies do not, on this evidence, materially shift the level at which an AI rater scores the same outputs.

\subsection{Stratified Scoring Protocol Effect}
\label{app:numerical-strat}

This sub-section provides the stratified analysis of the protocol $\times$ CDSS prompt configuration interaction---the dominant interaction identified in \S\ref{sec:results-cdss} (Figure~\ref{fig:heatmap})---selected for follow-up under the procedure of Appendix~\ref{app:lmm-strat}.

\begin{table}[!tbp]
\centering
\caption{Score distribution and protocol effect by evaluation question and CDSS prompt configuration stratum. For each question $\times$ stratum (Baseline, DRG), columns report the per-protocol mean (SD) and median (IQR) of AI rater scores, the GR $-$ Non-GR mean difference with standard error and 95\% Wald CI, and the $p$-value for the protocol effect within the stratum. Estimates are from the stratified LMM (Appendix~\ref{app:lmm-strat}). $n = 41{,}472$ observations per stratum per question.}
\label{tab:strat-protocol}
\includegraphics[width=\linewidth]{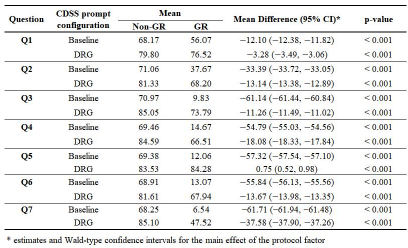}
\end{table}

Table~\ref{tab:strat-protocol} reports per-question Mean Difference estimates fit separately within each CDSS prompt configuration stratum. The scoring protocol effect within the Baseline stratum is consistently larger in magnitude than within the DRG stratum across all questions, except Q5, where the DRG stratum is near the upper end of the 0--100 scale under both scoring protocols (Mean Difference $= +0.75$, $p < 0.001$). The asymmetry reflects how the rubric interacts with each stratum's content. Under DRG the CDSS already received the patient-specific decision elements as reference documentation, so DRG outputs tend to address more rubric criteria explicitly than Baseline outputs. Applying GR therefore moves DRG scores less (DRG content already aligns with the rubric) and Baseline scores more (Baseline content does not). Q5 reverses the pattern only because its DRG cells already scored near the upper end of the 0--100 scale, leaving little room for a downward GR shift. The within-stratum Mean Difference estimates therefore measure how much each stratum's content already aligns with the rubric, not the scoring protocol effect itself.

\begin{figure}[!tbp]
\centering
\includegraphics[width=\linewidth]{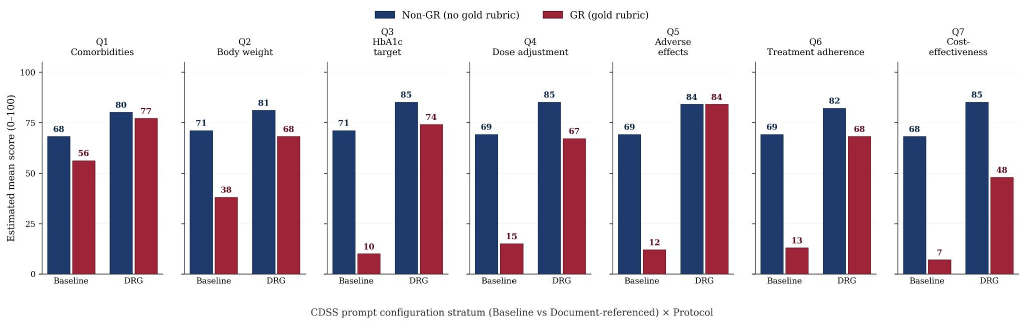}
\caption{Estimated mean score by evaluation question, CDSS prompt configuration stratum, and scoring protocol. Per-question bars from the stratified LMM (Appendix~\ref{app:lmm-strat})---four bars per question: Baseline$\times$Non-GR, Baseline$\times$GR, DRG$\times$Non-GR, DRG$\times$GR. Under Non-GR the DRG--Baseline gap is narrow and roughly constant across questions; under GR the gap widens substantially in every question except Q5, where DRG cells were already near the upper end of the 0--100 scale.}
\label{fig:stratified-bars}
\end{figure}

Figure~\ref{fig:stratified-bars} visualizes the same coefficients as a per-question bar plot. Three patterns are visible across the seven panels. Under Non-GR the DRG--Baseline gap is narrow and roughly constant across questions, indicating that the rubric-free scoring protocol does not vary its quality discrimination by clinical decision dimension. Under GR the DRG--Baseline gap widens substantially in every question except Q5---the same question where most DRG cells already scored near the upper end of the scale. The widening is not uniform: questions with more patient-specific rubric criteria show the largest within-stratum separation. For an AI rater pipeline that aims to rank CDSS outputs by quality, this is the visual signature of how rubric anchoring converts content-level differences in CDSS outputs into measurable score differences.

\subsection{Per-Level Decomposition of Protocol $\times$ Factor Interactions}
\label{app:numerical-decomp}

\begin{figure}[!tbp]
\centering
\includegraphics[width=\linewidth]{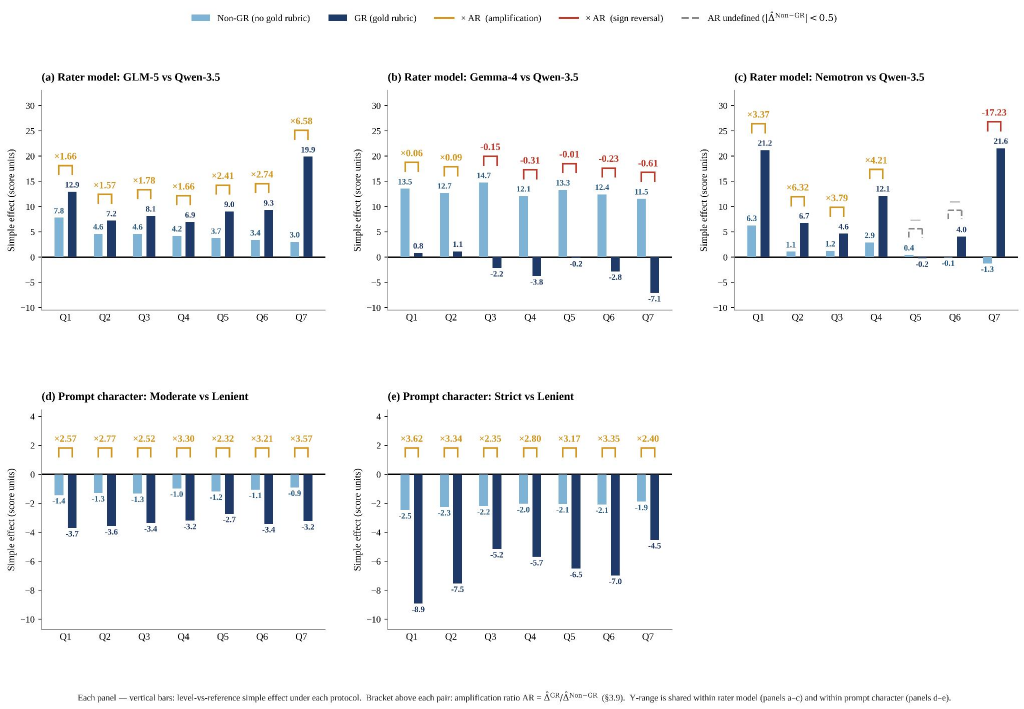}
\caption{Per-level decomposition of rater-side protocol $\times$ factor interactions. Five panels in two rows: top---(a)--(c) rater model (GLM-5, Gemma-4, Nemotron each vs.\ Qwen-3.5); bottom---(d)--(e) prompt character (Moderate, Strict each vs.\ Lenient). Each panel shows per-question simple effects (level vs.\ reference) under Non-GR (navy) and GR (maroon). The bracket above each bar pair reports the per-question amplification ratio (AR; Appendix~\ref{app:lmm-strat}), color-coded amber for amplification, red for sign reversal, and gray dashed when AR is in the unstable regime defined in Appendix~\ref{app:lmm-strat}. Y-range is shared within rater model (a--c) and within prompt character (d--e). GR uniformly amplifies prompt-character contrasts across every rater and question, but reshapes rater-model contrasts unevenly---including sign reversal for Gemma-4 vs.\ Qwen-3.5 across Q3--Q7. The protocol $\times$ prompt type interaction is omitted because every cell falls in the unstable regime; see Figure~\ref{fig:heatmap} and Figure~\ref{fig:prompt-type}.}
\label{fig:rater-decomp}
\end{figure}

\begin{figure}[!tbp]
\centering
\includegraphics[width=0.85\linewidth]{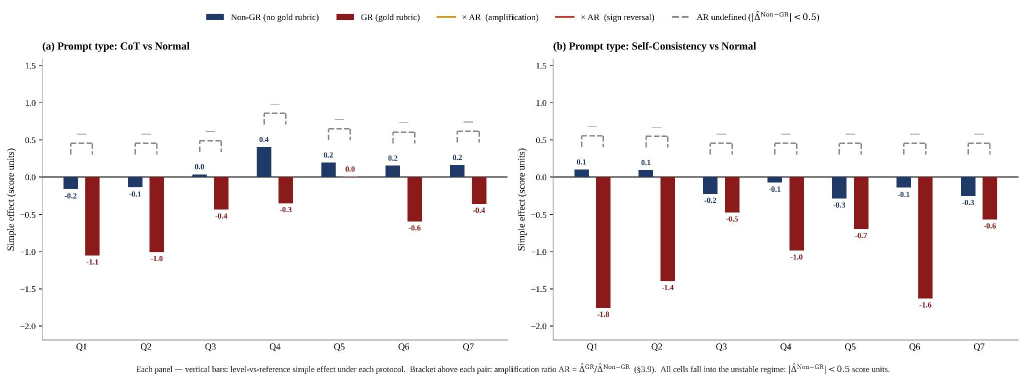}
\caption{Per-level decomposition of the protocol $\times$ prompt type interaction. Two panels: (a) CoT vs.\ Normal; (b) Self-Consistency vs.\ Normal. Each panel shows per-question simple effects (level vs.\ reference) under Non-GR (navy) and GR (maroon); the bracket above each bar pair reports the per-question amplification ratio (AR; Appendix~\ref{app:lmm-strat}), all of which fall in the unstable regime (gray dashed). The protocol $\times$ prompt type interaction is the smallest in the design---every per-question contrast falls in the AR-undefined regime under both protocols, indicating that structured deliberation does not supply patient-specific scoring criteria the way a rubric does.}
\label{fig:prompt-type}
\end{figure}

The factor-level Type 3 Wald $\chi^2/\mathrm{df}$ statistics for all five protocol $\times$ factor interactions are visualized in body Figure~\ref{fig:heatmap} (heatmap, with row-level Fisher-combined $p$-values). The per-level decomposition of the CDSS-side interactions is visualized in body Figure~\ref{fig:cdss-decomp}. This sub-section provides per-level decompositions for the rater-side interactions: the rater model and prompt character contrasts (Figure~\ref{fig:rater-decomp}), and the protocol $\times$ prompt type contrast (Figure~\ref{fig:prompt-type}), which falls entirely within the AR-undefined regime and is therefore reported separately.

Figure~\ref{fig:prompt-type} confirms what the factor-level Type 3 Wald result already implied (Figure~\ref{fig:heatmap}): the protocol $\times$ prompt type interaction is the smallest in the design and operates entirely within the AR-unstable regime. Every CoT--Normal and Self-Consistency--Normal contrast under Non-GR sits below 0.5 score units in absolute magnitude, leaving the AR numerically undefined despite the Fisher-combined per-question $p$-values reaching significance for several questions. The pattern is consistent with the \S\ref{sec:results-rater} interpretation: structured deliberation reorganizes how the rater applies its existing knowledge but does not supply patient-specific scoring criteria the way a rubric does.

\subsection{Random-Effect Variance Components}
\label{app:numerical-random}

\begin{table}[!tbp]
\centering
\caption{Random-effect variance components by evaluation question. Per-question variance estimates from the pooled LMM for the patient, prompt-order, and runs random intercepts and for the residual, reported in score-unit\textsuperscript{2} and as a percentage of total variance.}
\label{tab:re-variance}
\includegraphics[width=\linewidth]{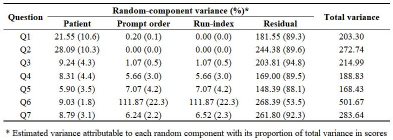}
\end{table}

Table~\ref{tab:re-variance} reports the four random-effect components per question, expressed as percentages of total variance. Q1--Q5 and Q7 are residual-dominated (88--95\%), with the patient component accounting for 3.1--10.6\%. Q6 is the only question where the prompt-order and runs components each carry substantial mass ($\sim 22\%$); the pattern indicates that the data did not separate the two cleanly at Q6, which also carries roughly twice the residual variance of Q3--Q5. Fixed-effect estimates on Q6 are unaffected. With the patient component below 11\% across six of seven questions, the sixteen patients function as approximately interchangeable evaluation contexts; the protocol-induced shift in Table~\ref{tab:score-dist} is therefore not driven by a small subset of unusual cases, and the scoring protocol contribution can be isolated without per-patient stratification.

Random-effect variance components were also compared between the two scoring protocols to characterize whether the protocol manipulation reshapes the clustering structure of AI rater scores beyond its effect on score levels. All three random-effect dimensions are summarized in score-point standard deviation units to enable direct comparison. The patient and run components moved in the direction the score-distribution result in \S\ref{sec:results-dist} would suggest: the across-patient SD increased from a mean of 2.87 (Non-GR) to 6.31 (GR), and the run-to-run SD increased from 5.79 to 9.08, indicating that GR scoring expands both between-patient heterogeneity and run-to-run variability. The order component moved in the opposite direction, with the across-order SD decreasing from a mean of 0.85 (Non-GR) to 0.48 (GR).

Interpretation of these patterns requires caution. The expanded GR-side variability in the patient and run components may reflect the rubric exposing patient-specific decision elements that the rater evaluates separately for each patient and at each run, but it may equivalently reflect the narrower Non-GR score band described in \S\ref{sec:results-dist}: when AI raters concentrate Non-GR scores within the 70--80 range regardless of patient or run, all three SD measures are mechanically compressed under Non-GR. The decrease in the order SD under GR is also consistent with two readings: the rubric may serve as a stable evaluative reference that attenuates sensitivity to surface prompt structure, or the order coefficients may simply remain small in magnitude relative to the overall GR-side variance. Distinguishing these explanations would require qualitative analysis of how the rater attends to each input component under the two protocols, which the present quantitative design does not support; we therefore report the variance pattern as observed and identify mechanistic interpretation as a target for follow-up work.

%==================================================================
\section{Rater Model Behavioral Variation (Full Analysis)}
\label{app:rater-defs}

This appendix provides full per-rater records for the five behavioral dimensions of AI rater behavior under each scoring protocol. The overall profile is shown in Figure~\ref{fig:perrater}; per-dimension numerical detail is provided in \S\ref{app:rater-mean}--\S\ref{app:rater-self} below. This appendix expands body \S\ref{sec:results-perrater}.

\begin{figure}[!tbp]
\centering
\includegraphics[width=\linewidth]{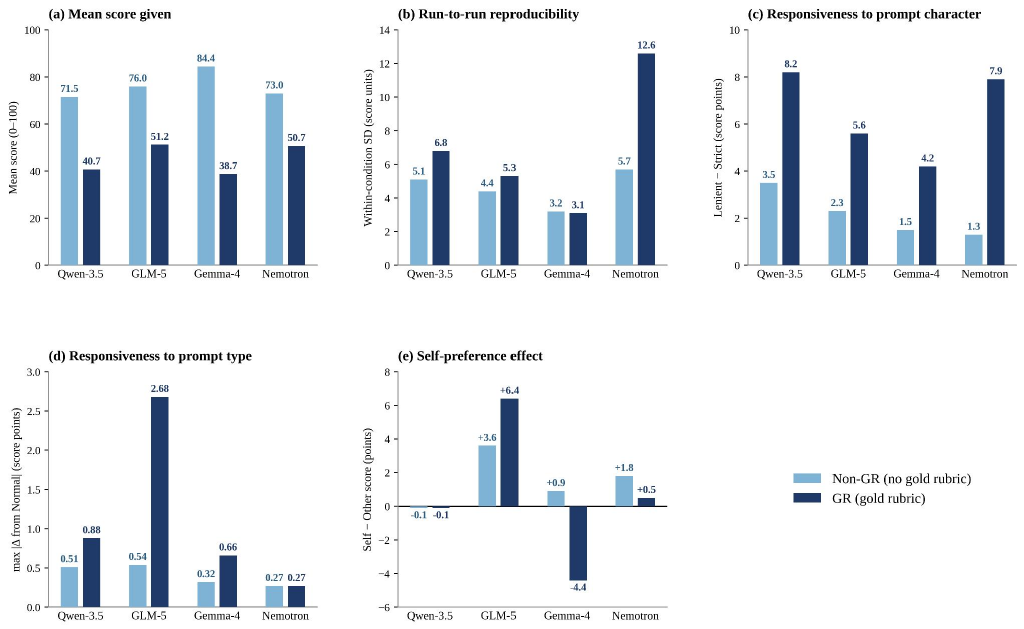}
\caption{Per-rater behavioral profile under each scoring protocol. Five panels (a--e) compare the four LLMs in their AI rater role (Qwen-3.5, GLM-5, Gemma-4, Nemotron) along five behavioral dimensions: (a) mean score, (b) run-to-run reproducibility, (c) responsiveness to prompt character, (d) responsiveness to prompt type, (e) self-preference effect. Each rater is shown with two bars per dimension---Non-GR in light blue, GR in dark navy. All five behavioral dimensions are rater-specific and protocol-conditional; per-rater protocol-induced shifts substantially varied (22.30--45.72 points), large enough to invalidate cross-protocol rater rankings. Operational definitions: Appendix~\ref{app:glossary}; full numerical records: \S\ref{app:rater-mean}--\S\ref{app:rater-self}.}
\label{fig:perrater}
\end{figure}

\subsection{Mean Score and Run-to-Run Reproducibility}
\label{app:rater-mean}

\begin{table}[!tbp]
\centering
\caption{Per-rater mean score and protocol-induced shift. Estimated mean score (0--100 scale) by rater $\times$ scoring protocol, the GR $-$ Non-GR shift with standard error, 95\% Wald CI, and $p$-value from the protocol $\times$ rater model interaction. $n = 10{,}368$ ratings per rater $\times$ protocol cell.}
\label{tab:rater-shift}
\includegraphics[width=\linewidth]{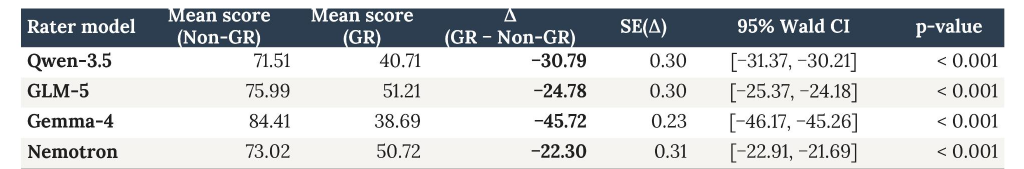}
\end{table}

Table~\ref{tab:rater-shift} expands the mean score dimension visualized in Figure~\ref{fig:perrater}, panel (a). The per-rater protocol-induced shifts span a factor of two: Gemma-4 moves from 84.41 to 38.69 (a 45.72-point shift, the largest in the design), Qwen-3.5 from 71.51 to 40.71 (30.79 points), GLM-5 from 75.99 to 51.21 (24.78 points), and Nemotron from 73.02 to 50.72 (22.30 points). All four per-rater shifts reach $p < 0.001$. For three of four raters in this design the per-rater shift is comparable to or larger than the per-question scoring protocol main effect itself, so a rater ranking obtained under one scoring protocol cannot be assumed to transfer to the other.

\subsection{Rater $\times$ Prompt Character (Responsiveness to Prompt Character)}
\label{app:rater-char}

\begin{table}[!tbp]
\centering
\caption{Per-rater mean score by prompt character $\times$ scoring protocol. Mean score at each prompt character level (Lenient, Moderate, Strict) by rater $\times$ protocol, the responsiveness to prompt character (Lenient $-$ Strict gap; Appendix~\ref{app:lmm-strat}), and the $p$-value for the per-rater Lenient $-$ Strict gap from the protocol $\times$ prompt character interaction.}
\label{tab:rater-character}
\includegraphics[width=\linewidth]{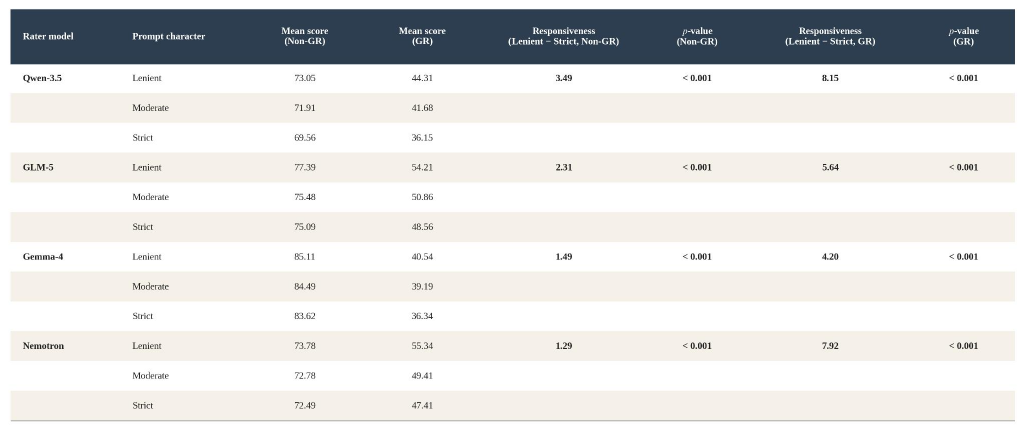}
\end{table}

\begin{figure}[!tbp]
\centering
\includegraphics[width=0.85\linewidth]{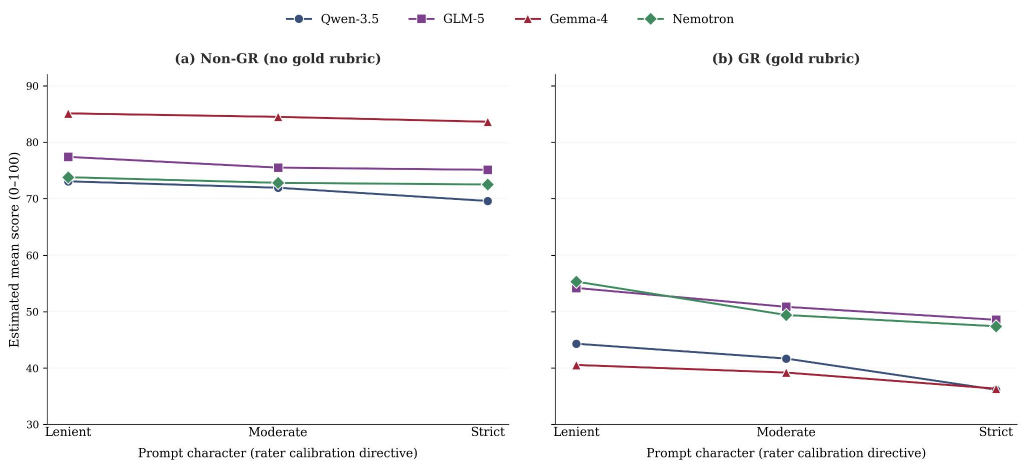}
\caption{Per-rater mean score across prompt character levels under each scoring protocol. Line plot of mean score by AI rater (Qwen-3.5, GLM-5, Gemma-4, Nemotron) across the three prompt character levels (Lenient, Moderate, Strict) under (a) Non-GR and (b) GR. The Lenient--Strict score gap amplifies under GR for every rater (Non-GR 1.3--3.5 points to GR 4.2--8.2 points), indicating that prompt-level calibration directives complement rather than substitute for rubric anchoring.}
\label{fig:rater-character}
\end{figure}

Table~\ref{tab:rater-character} reports the lenient/moderate/strict mean scores per rater $\times$ scoring protocol with the responsiveness to prompt character summary (Lenient $-$ Strict); Figure~\ref{fig:rater-character} visualizes the same data as line plots. The Lenient--Strict score gap amplifies under GR for every rater: Qwen-3.5 widens from 3.5 (Non-GR) to 8.2 (GR) score points, GLM-5 from 2.3 to 5.6, Gemma-4 from 1.5 to 4.2, and Nemotron from 1.3 to 7.9. Under Non-GR the prompt character directive shifts the mean by less than 4 points for every rater; under GR the same directive produces 4.2 to 8.2 point shifts. The pattern is interpretable: a calibration directive (`be strict') provides nothing for the rater to be strict about unless a rubric is present. With the rubric in place, the directive determines how leniently or strictly the rater applies each rubric criterion. Prompt-level calibration therefore complements rubric anchoring rather than substituting for it---a property that holds across all four raters in this design.

\subsection{Rater $\times$ Prompt Type (Responsiveness to Prompt Type)}
\label{app:rater-type-detailed}

\begin{table}[!tbp]
\centering
\caption{Per-rater mean score by prompt type $\times$ scoring protocol. Mean score at each prompt type level (Normal, CoT, SC) by rater $\times$ protocol, the responsiveness to prompt type (max $|\Delta$ from Normal$|$; Appendix~\ref{app:lmm-strat}), and the $p$-value for the per-rater max $|\Delta|$ from the protocol $\times$ prompt type interaction.}
\label{tab:rater-type}
\includegraphics[width=\linewidth]{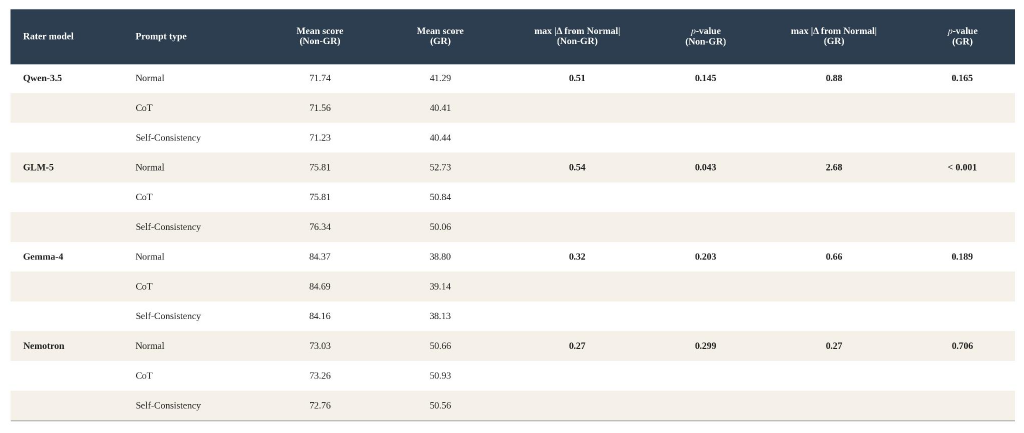}
\end{table}

\begin{figure}[!tbp]
\centering
\includegraphics[width=0.85\linewidth]{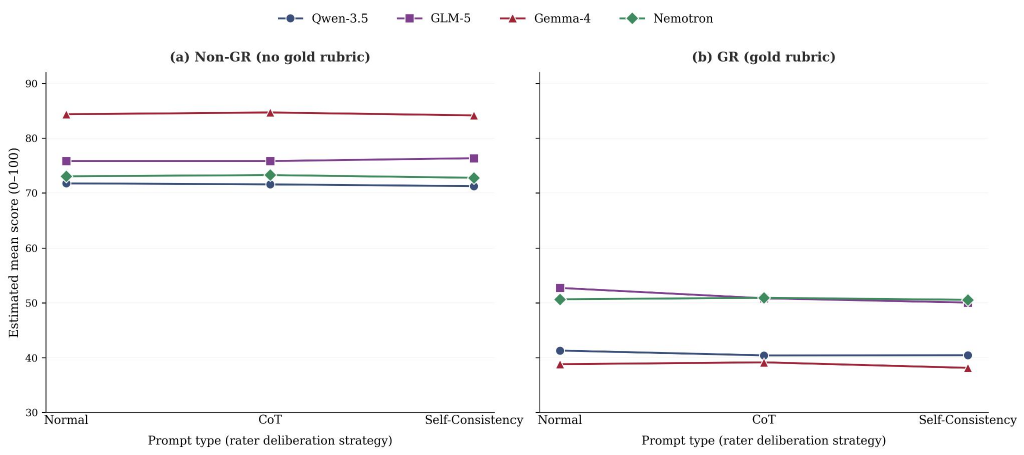}
\caption{Per-rater mean score across prompt type levels under each scoring protocol. Line plot of mean score by AI rater across the three prompt type levels (Normal, CoT, Self-Consistency) under (a) Non-GR and (b) GR; rater marker mapping matches Figure~\ref{fig:rater-character}. Prompt type effects remain within $\pm 2.7$ points across all rater $\times$ protocol cells (and within $\pm 1.0$ in seven of eight cells), an order of magnitude smaller than the prompt-character effects shown in Figure~\ref{fig:rater-character}.}
\label{fig:rater-type-fig}
\end{figure}

The prompt-type effects are an order of magnitude smaller than the prompt-character effects from Table~\ref{tab:rater-character} and Figure~\ref{fig:rater-character}. The responsiveness to prompt character dimension produces 1.3--8.2 point shifts depending on scoring protocol; the responsiveness to prompt type dimension produces shifts within $\pm 2.7$ points across all rater $\times$ scoring protocol cells, and within $\pm 1.0$ in seven of eight cells. Chain-of-Thought reasoning and Self-Consistency aggregation are well-documented to improve generation quality on reasoning tasks. The same strategies, applied on the rater side, do not move where a rater settles its score. An AI rater pipeline therefore cannot recover what rubric anchoring provides by routing scoring through CoT or Self-Consistency reasoning blocks: whatever benefit those blocks bring on the generator side does not transfer to the evaluator role.

\subsection{Run-to-Run Reproducibility}
\label{app:rater-reprod}

\begin{table}[!tbp]
\centering
\caption{Per-rater run-to-run reproducibility. Standard deviation (SD) and coefficient of variation (CV) by rater $\times$ scoring protocol, the SD ratio (SD$_\text{GR}$ $\div$ SD$_\text{Non-GR}$), and the $p$-value from Levene's test (Brown--Forsythe variant) for equality of variances across the two protocols. $n = 10{,}368$ ratings per cell.}
\label{tab:rater-reprod}
\includegraphics[width=\linewidth]{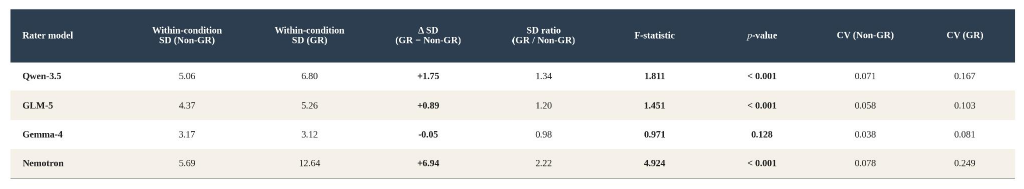}
\end{table}

Table~\ref{tab:rater-reprod} expands the run-to-run reproducibility dimension visualized in Figure~\ref{fig:perrater}, panel (b). Run-to-run stability is rater-specific and scoring-protocol-conditional. Gemma-4's standard deviation is 3.17 (Non-GR) and 3.12 (GR)---essentially unchanged. Nemotron's doubles from 5.69 to 12.64. The scoring protocol therefore does not just shift the rater's level; for some raters it also reshapes scoring stability---a property that is invisible from any single-protocol benchmark.

\subsection{Cross-Role 4$\times$4 Matrix}
\label{app:rater-cross}

\begin{figure}[!tbp]
\centering
\includegraphics[width=0.85\linewidth]{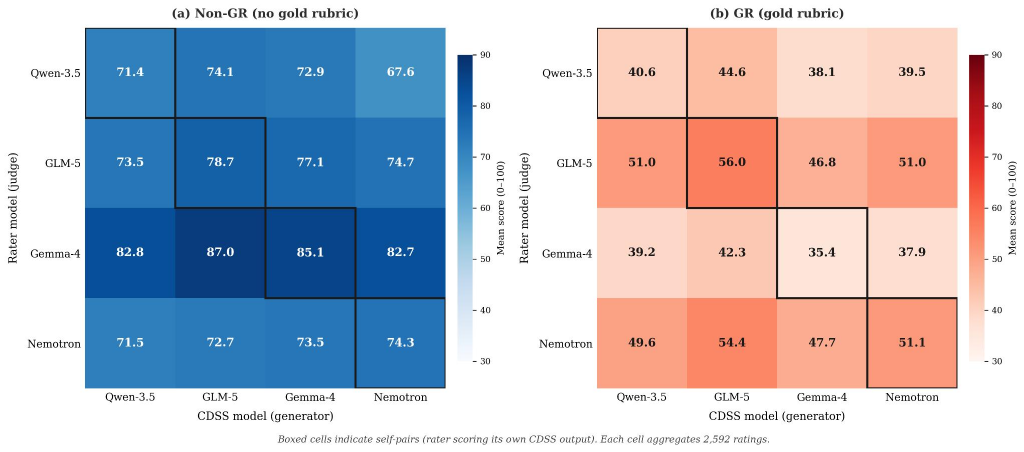}
\caption{Mean score by rater $\times$ CDSS model $\times$ scoring protocol. Two $4 \times 4$ heatmaps (Non-GR left, GR right): rows index the AI rater (judge); columns index the CDSS model (generator). Boxed cells mark self-pairs (rater scoring its own CDSS output). Each rater's row varies by less than 7 points across CDSS columns within each protocol, indicating that the per-rater protocol-induced shifts are properties of the rater itself rather than artifacts of specific rater--CDSS pairings. $n = 2{,}592$ ratings per cell.}
\label{fig:cross-role}
\end{figure}

Figure~\ref{fig:cross-role} isolates whether a rater's scoring behavior depends on whose CDSS output it is scoring. The most pronounced pattern is the Gemma-4 row: every cell $\geq 82.7$ under Non-GR (range 82.7--87.0) and every cell $\leq 42.3$ under GR (range 35.4--42.3). The near-uniformity within Gemma-4's row across all four CDSS columns shows that Gemma-4's lenient-then-strict reversal across protocols is a property of Gemma-4 itself, not of how it reads any particular CDSS model's output style. The same robustness applies to the other rows: each rater's row varies by less than 7 points across CDSS columns under Non-GR and less than 7 points under GR, indicating that the source of the CDSS output contributes only a small share of score variation once rater identity is fixed. For pipeline configuration this means the per-rater shifts in Table~\ref{tab:rater-shift} generalize across which CDSS the pipeline is evaluating---they are not artifacts of specific rater--CDSS combinations.

\subsection{Self-Preference Effect}
\label{app:rater-self}

\begin{table}[!tbp]
\centering
\caption{Self-preference effect by rater $\times$ scoring protocol. Mean self-rating, mean other-rating, self-preference effect (self $-$ other) with 95\% Wald CI, and $p$-value, by rater $\times$ protocol. $n$ is reported per cell in the table.}
\label{tab:self-pref}
\includegraphics[width=\linewidth]{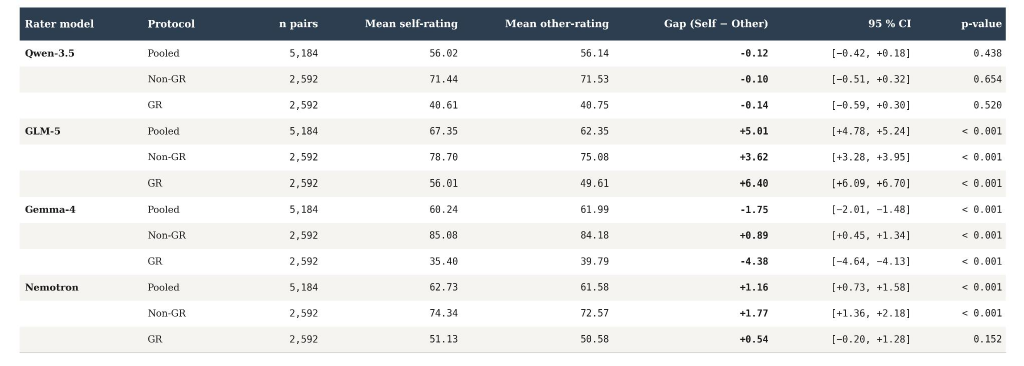}
\end{table}

Table~\ref{tab:self-pref} reports the self-preference effect with full inferential detail under Pooled, Non-GR, and GR scoring protocols. GLM-5 maintains a positive self-preference effect on both scoring protocols ($+3.62$ Non-GR; $+6.40$ GR). Gemma-4 reverses sign across protocols ($+0.89$ Non-GR; $-4.38$ GR). Qwen-3.5 is approximately self-neutral on both scoring protocols (effects within $\pm 0.15$). Nemotron's positive Non-GR effect ($+1.77$) is reduced to $+0.54$ under GR ($p = 0.152$) and is the only Nemotron self-preference comparison that does not reach significance. Self-preference has been reported as a stable rater-side bias in general-domain LLM-as-a-judge studies [8], with the implication that a model used as both generator and judge will systematically over-score its own outputs. The pattern here contradicts that assumption. GLM-5 fits the prior characterization, but Gemma-4 reverses sign across protocols---over-scoring its own outputs under Non-GR and under-scoring them under GR, a 5.27-point swing with both protocol-specific effects significant at $p < 0.001$. This pattern indicates that Gemma-4 applies the rubric more critically to its own outputs than to others'. For pipelines that use the same model as generator and judge, this means self-preference cannot be benchmarked once and assumed stable across scoring protocols.

%==================================================================
\section{Robustness Diagnostics}
\label{app:robust}

\subsection{Position Bias}
\label{app:robust-pos}

\begin{figure}[!tbp]
\centering
\includegraphics[width=\linewidth]{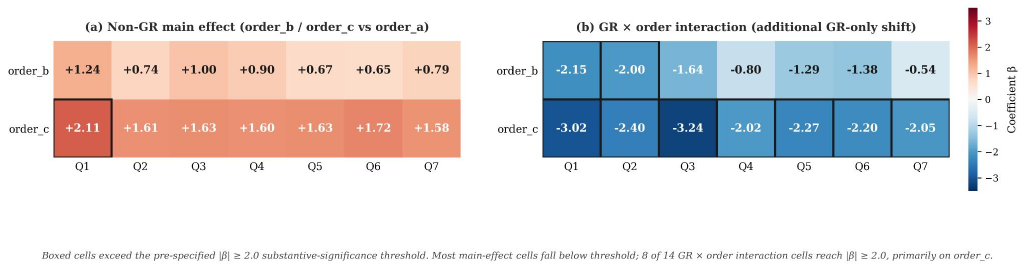}
\caption{Position-bias diagnostics for prompt order. Two heatmaps of per-question $\tau$ estimates: (a) Non-GR main effect (Order B vs.\ A; Order C vs.\ A); (b) protocol $\times$ prompt-order interaction. Boxed cells exceed the pre-specified threshold $|\tau| \geq 2.0$. Prompt-section ordering is negligible under Non-GR (panel a, largest $|\tau| = 2.11$); under GR a small but controllable order interaction emerges (panel b, largest $|\tau| = 3.24$ at Q3 $\times$ Order C), concentrated in Order C, which places the rubric-related sections last.}
\label{fig:position}
\end{figure}

Figure~\ref{fig:position} reports per-question $\tau$ estimates for prompt-order effects. The two-panel split is the diagnostic structure. Panel (a) asks whether the order in which the rater receives the prompt sections systematically biases scores in the absence of any scoring protocol manipulation; the answer is no---the largest $|\tau|$ in panel (a) is 2.11, marginal at the $|\tau| \geq 2.0$ threshold, and all other main-effect cells fall below threshold. Panel (b) asks whether the scoring protocol manipulation itself becomes order-sensitive, and the picture is different: 8 of 14 cells exceed the threshold, with Order C (the order that places the rubric-related sections last) carrying the larger interaction magnitudes and the maximum $|\tau| = 3.24$ occurring at (Q3, GR $\times$ Order C). GR scoring is therefore not as insensitive to prompt order as Non-GR scoring is. The robustness conclusion is scoring-protocol-conditional: position bias is negligible on the Non-GR side but enters as a small, controllable effect on the GR side. An AI rater pipeline that uses rubric anchoring should fix the prompt-section order across runs to keep scores comparable.

\end{document}